%% file: main.tex
\documentclass{article}
\usepackage{graphicx}
\usepackage{amsmath}
\usepackage{amssymb}
\usepackage{subcaption}
\usepackage{babel,blindtext}
\usepackage{booktabs}
\usepackage{multirow}
\usepackage{bm}
\usepackage{bbm}
\usepackage{xspace}
\usepackage{rotating}

\usepackage[final]{corl_2021} 
\usepackage{todonotes}

\newcommand{\methodname}[1]{tSLAM}
\newcommand{\vpad}[1]{-5mm}

\title{Curiosity Driven Self-supervised Tactile Exploration of Unknown Objects}

\author{
  Jianren Wang \thanks{Contribute equally.}\\
  Carnegie Mellon University \\
  \texttt{jianrenw@andrew.cmu.edu} \\
  \And
  Yujie Lu\textsuperscript{\textasteriskcentered}\\
  UC Santa Barbara \\
  \texttt{yujielu@ucsb.edu} \\
  \And
  Vikash Kumar\\
  Facebook AI Research\\
  \texttt{vikash@cs.washington.edu} \\
}

\begin{document}
\maketitle


\begin{abstract}
    Intricate behaviors an organism can exhibit is predicated on its ability to sense and effectively interpret complexities of its surroundings. Relevant information is often distributed between multiple modalities, and requires the organism to exhibit information assimilation capabilities in addition to information seeking behaviors. While biological beings leverage multiple sensing modalities for decision making, current robots are overly reliant on visual inputs. In this work, we want to augment our robots with the ability to leverage the (relatively under-explored) modality of touch. To focus our investigation, we study the problem of scene reconstruction where touch is the only available sensing modality. We present Tactile Slam (\methodname~) -- which prepares an agent to acquire information seeking behavior and use implicit understanding of common household items to reconstruct the geometric details of the object under exploration. Using the anthropomorphic `ADROIT' hand, we demonstrate that \methodname~ is highly effective in reconstructing objects of varying complexities within 6 seconds of interactions. We also established the generality of \methodname~ by training only on 3D Warehouse objects and testing on ContactDB objects.
    Please refer to https://sites.google.com/view/tslam for more visualization.
\end{abstract}

\keywords{Self-supervised Exploration, Tactile Sensing, 3D Reconstruction} 


\input{sections/1-introduction.tex}

\input{sections/2-related_works.tex}

\input{sections/3-preliminaries.tex}

\input{sections/4-methods.tex}

\input{sections/5-experiments.tex}

\input{sections/6-conclusion.tex}


\clearpage
\acknowledgments{If a paper is accepted, the final camera-ready version will (and probably should) include acknowledgments. All acknowledgments go at the end of the paper, including thanks to reviewers who gave useful comments, to colleagues who contributed to the ideas, and to funding agencies and corporate sponsors that provided financial support.}


\bibliography{reference}  

\end{document}

%% file: sections/1-introduction.tex
\section{Introduction}


Biological beings have continuously evolved to maintain a competitive edge in their surroundings. As complexities and competition in the environment evolved, so did the need for complex decisions and intelligent behavior. Not only did the diversity of sensors explode (eyes, ears, fins, whiskers, etc.), organisms relied on an increasing larger number of sensory inputs (light, sound, pressure, electrostatics, etc.) for their behaviors~\cite{romei2012sounds,warren2002perception}. While visual and auditory modalities became common in most, primates significantly evolved their sense of touch as they started aggressively manipulating their environments to their advantage~\cite{klatzky1985identifying}. We hypothesize that the current state of robotics is at similar crossroads. While visual and proprioceptive inputs are common, the modality of touch (as well as acoustics) are relatively underexplored and needs further attention.

Unlike visual stimulation, tactile inputs do not require a direct line of sight. Their strengths are quite complementary to each other. While vision brings relevant information during search and broad localization, it completely fails during manipulation due to the occlusion. This is precisely when the modality of touch gets activated. While former (visual search~\cite{renfaster, tung2019learning} and visual servoing~\cite{levine2016end, agrawal2016learning}) has found significant attention, the later has been relatively underexplored.

\begin{figure}[!h]
\centering
\includegraphics[width=0.8\textwidth]{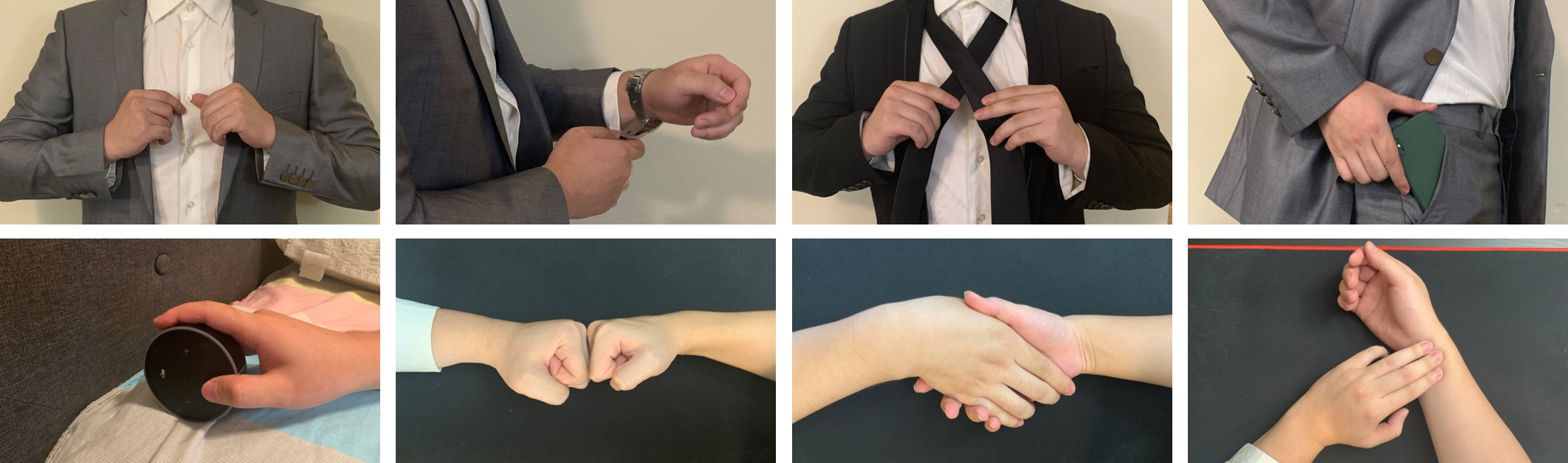}
\caption{Humans seek information from tactile-only sensing in many daily activities, social activities, and health applications. }
\label{fig:teaser}
\end{figure}
This work is primarily focused on enabling our robotic agents to attention to the modality of touch. To focus our investigation, we concentrate on regimes where interaction has been localized to the area of attention and no other modality but touch is present to seek information. While this might seems a bit contrived, we were surprised by the prevalence of this setting (Fig.~\ref{fig:teaser}) in (a) activities of daily living -- buttoning / unbuttoning shirt, wearing a wrist watch, taking things out of pocket, tying/ untying a knot, searching for object in the dark, etc. (b) social settings -- hand shakes, hugs, affectionate behaviors, etc. as well as (c) health and medial applications -- checking pulse, breast and prostrate cancer checks, massages etc.

Tactile signals contains high frequency but local interaction information. Most prior works leveraging the modality of touch ~\cite{sommer2014bimanual, pezzementi2011object, bierbaum2008robust,ottenhaus2016local} have focused on fusing the visual and touch features. The local information provided by touch are registered and localized against the global features provided by visual inputs. In contrast, as noted above, there are abundance of setting where tactile is sole modality available and has no additional information for global registration. Owing to the complexity of the problem, such settings has been largely overlooked and have hindered progress in robotic manipulation skills. 

On the other hand, humans can efficiently seek information and utilize the gathered information with real world priors to solve task blindfolded using only tactile sensation. Effective understanding of tactile modality under such settings requires (a) Information seeking behaviors, as well as (b) information assimilation capabilities that can consolidate information spread temporally as well as spatially. In this work, we ground these challenges in the context of simultaneous localization and mapping (SLAM) using touch as the only sensing modality. 

\textbf{Our contributions}
In this paper, we develop \methodname~, 
\begin{enumerate}
    \item A self supervised methods that leverages curiosity based objectives to impart information seeking behaviors to a tactile agent.
    \item \methodname~ leverages implicit understanding of shapes acquired using common household object to accelerate reconstructions.
    \item Can deliver detailed geometric features of unseen objects with varying levels of complexity within 6 seconds of interaction.
    \item \methodname~ is effective even for convex objects and for objects with large voids.
\end{enumerate}




%% file: sections/2-related_works.tex
\section{Related Works}

\subsection{Tactile sensors}
A large range of technologies are used for tactile sensing. A class of sensors that has recently proved versatile is vision-based sensors, which measures contact forces as changes in images recorded by a camera. Previous vision-based tactile sensors include TacTip~\cite{ward2018tactip}, FingerVision~\cite{yamaguchi2016combining}, GelSight~\cite{yuan2017gelsight}, DIGIT~\cite{lambeta2020digit}, and several others~\cite{shimonomura2019tactile}. However, these sensors are always complex and expensive. For simplicity, we use binary switch sensors in our setting. Contact switches permit the detection of discrete on/off events brought about by mechanical contact~\cite{webster1988tactile}. The ease of designing and building this type of sensor has permitted its integration into a wide variety of robotic systems~\cite{edin2006bio}.

\subsection{Tactile sensing}
While haptic signals have been exploited for efficient exploration~\cite{bierbaum2008potential,yi2016active,martinez2013active,jamali2016active,driess2017active} and shape completion~\cite{sommer2014bimanual, pezzementi2011object, bierbaum2008robust,ottenhaus2016local}, a general technique for information seeking and information assimilation still remains an open question. Suresh et al.~\cite{Suresh21tactile} combines efficient Gaussian process implicit surfaces (GPIS)~\cite{dragiev2011gaussian} regression with factor graph optimization for planar shape inference. However, it only works with simple pusher-slider under planar environment. Bierbaum et al.~\cite{bierbaum2008potential} presents a tactile exploration strategy to guide an anthropomorphic hand along the surface of previously unknown objects and build a 3D object representation based on acquired tactile point clouds. 

Tactile sensing also plays an important role in soft robotics~\cite{hakozaki1999telemetric, hoshi2006robot, yuan2019soft}. Yang et al.~\cite{yang2020scalable} shows soft robot finger can identify sectional diameters and structural strains at a very high accuracy. Thuruthl et al.~\cite{thuruthel2019soft} further shows soft robot integrated with tactile sensor can model force and deformation of soft robotic systems.

\subsection{Self-supervised Exploration}
Consider an agent that sees an observation, takes an action and transitions to the next state. We aim to incentivize this agent with a reward relating to how informative the transition was, so that the agent can explore the complicated environment more efficiently. 
One simple approach to encourage exploration is to use state visitation counts~\cite{bellemare2016unifying, fu2017ex2, tang2017exploration}, where one maximizes visits on less frequent states. However, counting in the continuous space is usually challenging.
Recently a more popular line of works are using prediction error~\cite{Schmidhuber:1991:PIC:116517.116542,burda2018large}, prediction uncertainty~\cite{houthooft2016vime, osband2016deep}, or improvement~\cite{lopes2012exploration} of a forward dynamics or value model as intrinsic rewards. As a result, the agent is driven to reach regions of the environment that are difficult to reason with the current model.

\subsection{Reconstructions}
There is a vast literature addressing 3D shape reconstruction from visual signals. \textit{e.g.} RGB image~\cite{smith2017improved},depth images~\cite{rock2015completing} and even thermal images~\cite{acampora20113d}. With recent development of implicit functions~\cite{xu2019disn}, these methods achieve high-fidelity reconstruction.
Some prior works leverage the modality of touch~\cite{allen1990sensordata, bierbaum2008haptic, takamitsu2017robotics}. More recent works also exploit vision and touch for 3D shape reconstruction~\cite{VisionTouch}.
 With the use of high-resolution sensors~\cite{lambeta2020digit}, these works consistently improves single modality baselines. However, using only touch to achieve high-fidelity reconstruction is largely unexplored. Few works attempt to explore the geometry of unseen objects with heuristic policies~\cite{yi2016active, sommer2014bimanual}. However, most reconstructions are sparse point clouds~\cite{bierbaum2008robust}, and planar contours~\cite{Suresh21tactile}. In contrast, we propose to use implicit functions that encode shape priors for a high-fidelity reconstruction.

%% file: sections/3-preliminaries.tex
\section{Problem definition}

\subsection{Tactile SLAM}
By Tactile SLAM, we refer to the problem of exploring and reconstructing the geometric details of an object solely using the modality of touch. We study the problem under the tabletop settings where the manipulator has already been localized near the object under investigation and no modality but touch is present to seek information. We assume time-synced binary information (touch/ no-touch) is available, and place no restriction on the number of tactile sensors. We rely on robot's forward kinematics (FK) to access the location of the contact point. 

Unlike cameras, tactile sensors are short range devices that only captures local information using interaction forces between the manipulator and the object. Acquiring global information about the object requires effective exploration, localization, and assimilation of information from multiple sensors capturing data over an extended period of time. We allowed our agent to interact with the object for a fixed amount of time before delivering a mesh with geometric details. The agent has access to enough memory to store and update implicit object representation in place however there is no extra memory to store interaction data.

\subsection{Learning to explore}
\label{sec:setting_learning}
Effective reconstruction requires agent to exhibit information gathering strategy based on tactile features of objects. We formulate such problem by using the standard Markov Decision Process (MDP)~\cite{bellman1957markovian} $\mathcal{M}$, with state space $\mathcal{S}$, action space $\mathcal{A}$, state transition dynamics $\mathcal{T}: \mathcal{S} \times \mathcal{A} \rightarrow \mathcal{S}$, reward function $r: \mathcal{S} \times \mathcal{A} \rightarrow R$, horizon $\mathcal{H}$, and discount factor $\gamma \in (0,1]$. To determine the optimal stochastic policy $\pi: \mathcal{S} \rightarrow \mathcal{\mathcal{A}}$, we need maximized expected discounted reward


\begin{equation}
J(\pi) = \mathbf{E}_\pi[\sum_0^{\mathcal{H}-1} \gamma r(s_t,a_t)]
\end{equation}

We represent the policy by a neural network with parameters $\theta_\pi$ that are learned as described in Sec.~\ref{sec:exploration}. $\mathcal{S}$ is defined in Sec.~\ref{sec:exploration}. $\mathcal{A}$, $\mathcal{H}$, $\gamma$ are defined in Sec.~\ref{sec:exp_setting}. Please refer to Supplementary Materials for more details.

\subsection{Learning to reconstruct}
Agent is provided access to a large corpus of general household items to acquire shape priors to aid and accelerate reconstructions. During inference, given an incomplete or low resolution 3D voxel $X = R^{N \times N \times N}$ ($N \in \mathcal{N}$ denotes the input resolution), and a 3D point $p \in R^3$, we want the agent to predict if p lies inside or outside the object.


%% file: sections/4-methods.tex
 \section{\methodname ~: Method Details}

\begin{figure}
  \centering
  \includegraphics[width=0.8\textwidth]{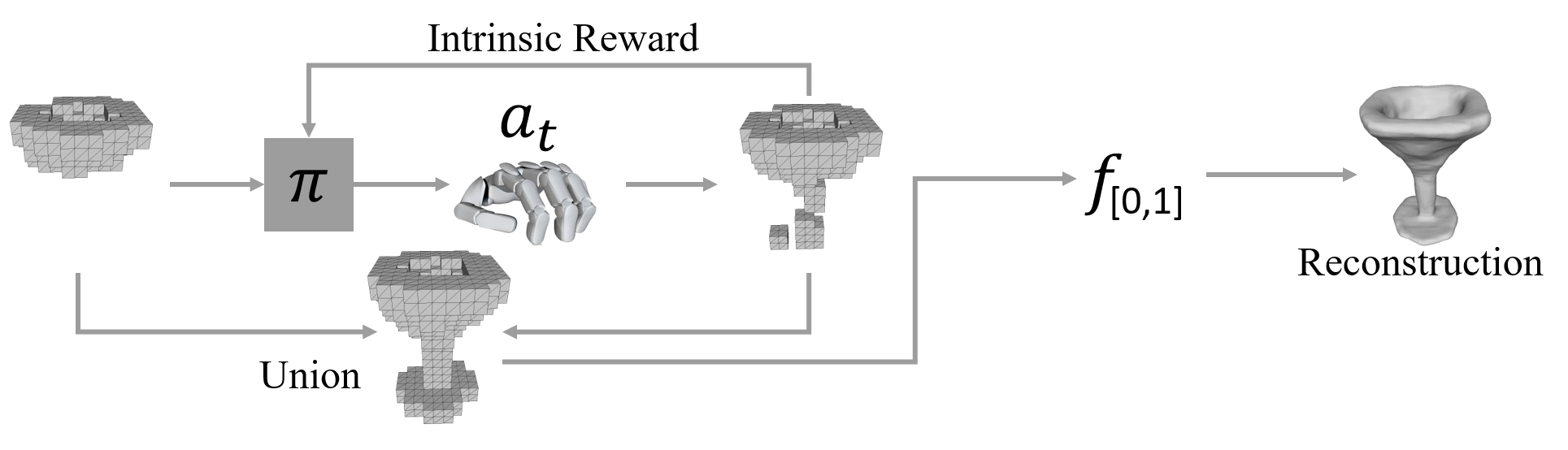}
  \caption{tSLAM Architecture: at time step $t$, an agent takes action $a_t$ given a occupancy grid observation $G^t$ and robot's joint angle sensors as inputs and ends up in a new state. The discovery of new occupancy grid is used as intrinsic rewards to train the policy $\pi$. After $\mathcal{H}$ steps, a union of all contact points are fed to an implicit function for detailed reconstruction.}
  \label{fig:pipeline}
\end{figure}

\methodname~ is composed of a self-supervised exploration policy that incentives agents to maximize the discovery of new contact points and the coverage of exploration space (Section~\ref{sec:exploration}), and a reconstruction model that recovers the underlying geometry of the objects using only tactile sensing (Section~\ref{sec:reconstruction}). We provide an overview of the method in Figure~\ref{fig:pipeline} before detailing each sub-module.

\subsection{Self-supervised Exploration}
\label{sec:exploration}

To make the hand interact with an object efficiently, we propose a self-supervised exploration policy that only requires binary touch as inputs. As mentioned in Sec.~\ref{sec:setting_learning}, we formulate the problem of effective exploration as a finite-horizon, discounted MDP, where $\mathcal{S}$ is represented by a volumetric occupancy grid~\cite{moravec1985high} and hand joint angles. Mathematically, the robot workspace is represented as a 3D occupancy grid $G = R^{N \times N \times N}$ ($N \in \mathcal{N}$ denotes the resolution). The value of each grid is either 0 or 1, depending on whether the grid is occupied by an object. By default, all unvisited grid will be set to zero at the beginning. A grid will be set to 1 if and only if it is visited by the agent and it is occupied by an object. 



For the reward function $r$, we can directly use the overlap of reconstructed and true object shape. But its not practical in the real world as we don't have access to true shape, it needs to be reconstructed. We leverage curiosity based intrinsic motivation to train an agent to exhibit effective tactile exploration strategies. The agent is incentived using novel part discovery $r_d$ and visitation count $r_c$ based objectives :

\begin{align}
    r_d = \sum_{i=1}^{N \times N \times N} \mathbbm{1}(H \cap G_i \in O)
\end{align}
where $\mathbbm{1}(H \cap G_i \in O) \in \{0,1\}$ indicates whether dexterous hand $H$ intersect with novel part of the object $O$. Please refer to Section~\ref{sec:results} for different choices and their performance.

Inspired by visitation count~\cite{bellemare2016unifying, lopes2012exploration}, we also propose an environment coverage reward $r_c$ in additional to the novel part discovery reward. As suggested, the environment coverage reward encourage the agent to explore novel states of the environment. At timestep $t$, the environment coverage reward is defined as:

\begin{align}
    r_c = \sum_{i=1}^{N \times N \times N} \mathbbm{1}(H \cap G_i \not\in G^{t-1})
\end{align}
where $G^{t-1}$ represents all visited grids until $t-1$, $\mathbbm{1}(H \cap G_i \not\in G^{t-1}) \in \{0,1\}$ indicates whether dexterous hand $H$ intersect with a novel occupancy grid $G_i$. Intuitively, the environment coverage reward will help the agent to get around local minima of state spaces, e.g. repeatedly visit a small region with complex texture. 

To summarize, we use both novel part discovery reward and environment coverage reward as intrinsic rewards $r^t = r^t_d + \lambda \times r^t_c$, where $\lambda$ is a weight factor. The agent is optimized using PPO~\cite{schulman2017proximal} to maximize the expected reward $J(\pi)$.


At the end of the horizon $\mathcal{H}$, all information collected are stored in the volumetric occupancy grid $G^\mathcal{H}$, which can be further used for reconstruction and other down-streaming tasks.

\subsection{3D Recontruction and Completion}
\label{sec:reconstruction}

For a better understanding of the underlying geometry of the objects, we propose to build a reconstruction model above the information collected during exploration-phase. The reconstruction model should encode a good knowledge of shape priors and should be able to reconstruct high-fidelity objects. We adopt IF-Nets~\cite{chibane2020implicit}, an implicit functions in feature space for 3D shape reconstruction and completion.

Specifically, given an incomplete or low resolution 3D voxel $X$, we compute a 3D grid of multi-scale features $F_1$,...,$F_n$, The feature grids $F_k$ at the early stages (starting at $k=1$) capture high frequencies (shape detail), whereas feature grids $F_k$ at the late stages (ending at stage k = n) have a large receptive fields, which capture the global structure of the data.

Instead of classifying point coordinates $p$ directly, we extract the learned deep features $F_1(p)$,...,$F_n(p)$ from the feature grids at location $p$. Since feature grids are discrete, we use trilinear interpolation to query continuous 3D points $p \in R^3$. The point encoding $F_1(p)$,...,$F_n(p)$, with $F_k(p) \in F_k$, is then fed into a point-wise decoder $f(\dot)$, parameterized by a fully connected neural network, to predict if the point $p$ lies inside or outside the shape:

\begin{align}
    f(F_1(p),...,F_n(p)): \mathcal{F}_1 \times ... \times \mathcal{F}_n \rightarrow [0,1]
\end{align}


While \methodname~ sequentially apply above two steps (exploration and reconstruction), it can be a monolithic piece by extending to an iterative process and achieve better outcomes. We discuss these in the Section~\ref{sec: future_works}.

%% file: sections/5-experiments.tex
\section{Experiments}

We evaluate the performance of \methodname~ using Adroit Manipulation Platform~\cite{Kumar2016thesis} with 60 objects from ContactDB~\cite{Brahmbhatt_2019_CVPR} dataset. We analyze the effectiveness of each component of our method by performing ablation analysis. In addition, we perform quantitative and qualitative experiments to validate effectiveness of \methodname~ in reconstructing unknown objects of varying complexities.

\subsection{Dataset and Experiment Setup}
\label{sec:exp_setting}

Adroit Manipulation Platform (Fig.~\ref{fig:trajectory}) is comprised of the Shadow Hand skeleton~\cite{walker2005shadow} and a custom arm powered using a custom high power low latency actuation system. Robot's action space is 28-dimensional: a) 3-dimensions specify the robot arm in Cartesian coordinates ($x,y,z$) b) 1-dimension specify the robot arm in Euler coordinates (roll) c) 2-dimensions specify the robot wrist in Euler coordinates (roll,pitch) c) 22-dimension specify the finger joints of hand. In this environment, we discretize the environment space into a $32 \times 32 \times 32$ occupancy grid. For object sets, we consider using ContactDB~\cite{Brahmbhatt_2019_CVPR}, a challenging contact maps prediction benchmark of household 3D objects to  evaluate the effectiveness of our proposed method. We focus our analysis on three well established metrics proposed by~\cite{mescheder2019occupancy}. First, \textit{Volumetric intersection over union (IoU)}, which is defined as the quotient of the volume of the two meshes’ union and their intersection. Second, \textit{$Chamfer-L_2$ distance}, which is the mean of accuracy and completeness metric measured based on the mean distance of points on output mesh to nearest neighbors on ground truth mesh. Third, \textit{Normal consistency} score, which is the mean absolute dot product of the normals in one mesh and the normals at the corresponding nearest neighbors in the GT mesh. 
To evaluate the generalizability of our proposed method, both tactile exploration policy and the implicit feature network are trained using 3D Warehouse~\cite{warehouse} and tested on ContactDB~\cite{Brahmbhatt_2019_CVPR}. The tactile policy is trained on 600 objects from 50 categories of 3D Warehouse. We used six 3D convolutional layers to extract voxel features. We use a 4-layer multi-layer perceptron (MLP) as our policy network and used PPO~\cite{schulman2017ppo} to maximized the intrinsic reward with an Adam Optimizer. Hand position is randomly initialized at the beginning of every interaction episode of length 200 steps. The policy is trained with a total budget of 10M environment interaction steps with $\gamma$ set as ${0.99}$. Please refer to Supplementary Materials for training details.

\subsection{Results}
\label{sec:results}

We present the final reconstruction results of \methodname~ in Figure~\ref{fig:main_results} and compare the performance of our method to the various baselines using the outlined metrics in Table~\ref{table:main_results}. We compare our method with two baselines.
\begin{itemize}
    \item \textit{Random Policy}: The policy randomly moves in the action space for tactile exploration.
    \item \textit{Heuristic Policy}: A heuristically designed policy that induces power grasp from a randomly initialized open position.
\end{itemize}


As show in Table~\ref{table:main_results}, our method ourperforms baselines by a large margin. During exploration stage, our method improve over the performance of Random Policy by 13.46\% IoU and Heuristic Policy by 6.00\% (Occupancy Grid). With better perception model, our method further improve over the performance of Random Policy (17.50\%) and Heuristic Policy (11.52\%) (Reconstruction). 

\begin{table}
\centering
\begin{tabular}{c|cc|cc|cc} 
& \multicolumn{2}{|c|}{$IoU\uparrow$} & \multicolumn{2}{|c|}{$Chamfer-L_2\downarrow$} & \multicolumn{2}{|c}{$Normal-Consis.\uparrow$}\\
\hline
Occupancy Grid (Random) & 0.2617 & 0.3307 & 0.1100 & 0.0558 & 0.4211 & 0.4797\\
Occupancy Grid (Heuristic) & 0.3363 & 0.3828 & 0.0696 & 0.0473 & 0.4993 & 0.5871\\
Occupancy Grid (tSLAM) & \textbf{0.3963} & \textbf{0.4049} & \textbf{0.0458} & \textbf{0.0454} & \textbf{0.6554} & \textbf{0.7276}\\
\hline
Reconstruction (Random) & 0.2361 & 0.3134 & 0.1959 & 0.0914 & 0.5399 & 0.6318\\
Reconstruction (Heuristic) & 0.3059 & 0.3653 & 0.1151 & 0.0627 & 0.6330 & 0.7310\\
Reconstruction (tSLAM) & \textbf{0.4111} & \textbf{0.4287} & \textbf{0.0418} & \textbf{0.0349} & \textbf{0.7912} & \textbf{0.8352}\\
\end{tabular}
\caption{Results of point cloud reconstruction on ContactDB. Left number indicates score from 4 poses, right one from 8 poses. Reconstruction is the output of 3D object reconstruction network with Occupancy Grid as input. Metric is averaged on 60 objects. $Chamfer-L_2$ results $\times 10^{-2}$.}
\label{table:main_results}
\end{table}

\begin{table}
\centering
\begin{tabular}{c|cc|cc|cc} 
& \multicolumn{2}{|c|}{$IoU\uparrow$} & \multicolumn{2}{|c|}{$Chamfer-L_2\downarrow$} & \multicolumn{2}{|c}{$Normal-Consis.\uparrow$}\\
\hline
Ours(-Coverage) & 0.3647 & 0.4037 & 0.0638 & 0.0409 & 0.7158 & 0.7919\\
Ours(-Curiosity) & 0.2882 & 0.3576 & 0.1267 & 0.0710 & 0.5964 & 0.6925\\
Ours(Knn) & 0.4071 & 0.4252 & 0.0472 & 0.0464 & 0.7888 & 0.8290\\
Ours(\# Points) & 0.3688 & 0.3997 & 0.0631 & 0.0412 & 0.7269 & 0.7935\\
Ours(\# Contact) & 0.2294 & 0.3167 & 0.1868 & 0.0921 & 0.5432 & 0.6414\\
Ours(Disagreement) & 0.4111 & 0.4257 & 0.0477 & 0.0372 & 0.7798 & 0.8247\\
Ours(Chamfer) & 0.3603 & 0.3986 & 0.0547 & 0.0451 & 0.7173 & 0.7945\\
\hline
Ours & \textbf{0.4111} & \textbf{0.4287} & \textbf{0.0418} & \textbf{0.0349} & \textbf{0.7912} & \textbf{0.8352}\\
\end{tabular}
\caption{Results of point cloud reconstruction on ContactDB. Left number indicates score from 4 poses, right one from 8 poses for each object. Metrics are averaged on 60 objects. $Chamfer-L_2$ results $\times 10^{-2}$.}
\label{table:ablation}
\end{table}

\begin{table}
\centering
\setlength\tabcolsep{0.3pt}
\begin{tabular}{cccccccccccc}
& \small{R} & \small{H} & \small{O(-Cov)} & \small{O(-Cur)} & \small{O(Knn)} & \small{O(Pts)} & \small{O(Cts)} & \small{O(Dis)} & \small{O(Ch)} & \small{Ours} & \small{GT}\\
{\begin{sideways} \centering \small{grid} \end{sideways}} &
\includegraphics[width=.085\linewidth]{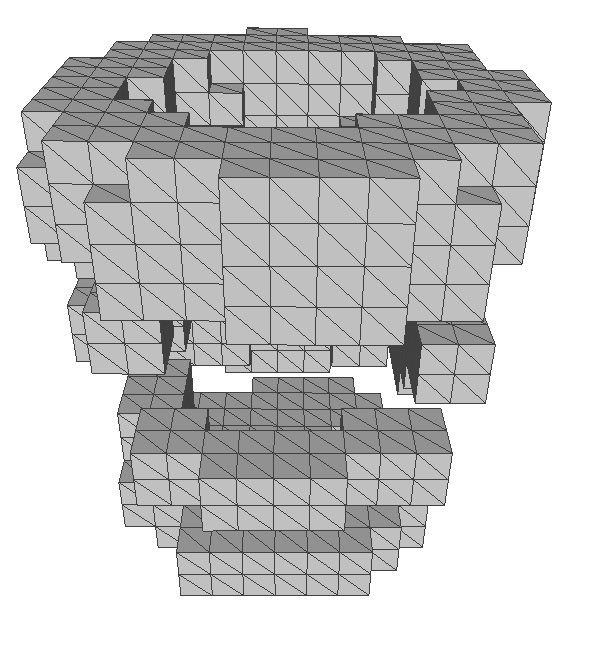} &
\includegraphics[width=.085\linewidth]{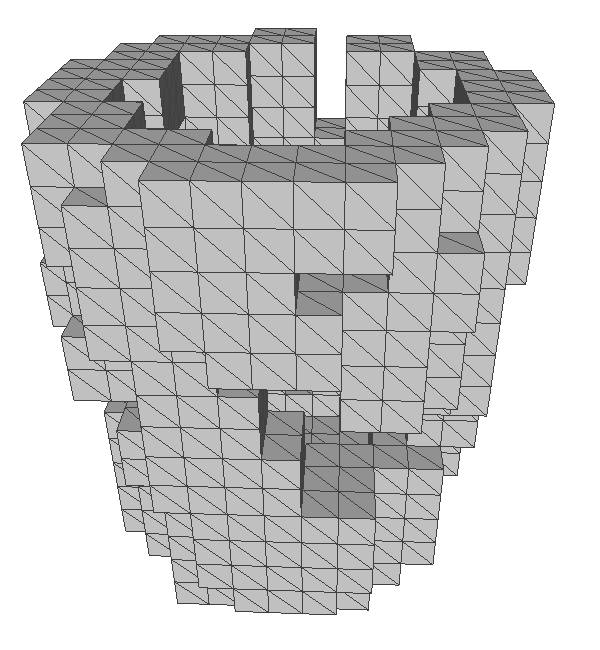} &
\includegraphics[width=.085\linewidth]{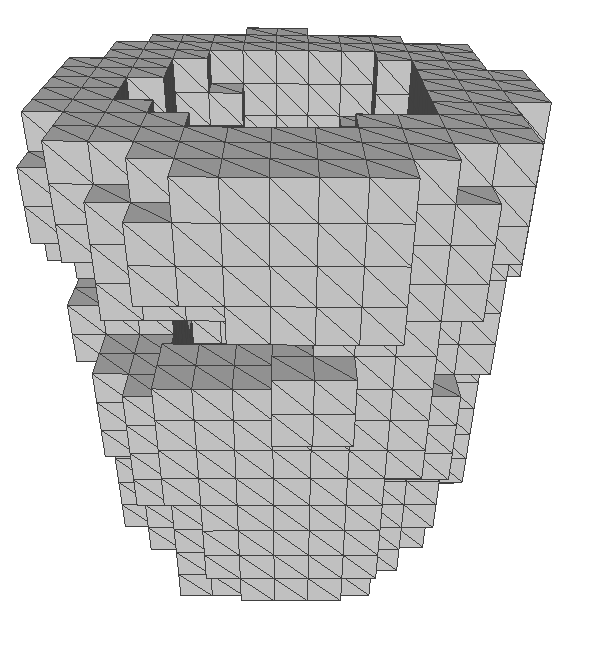} &
\includegraphics[width=.085\linewidth]{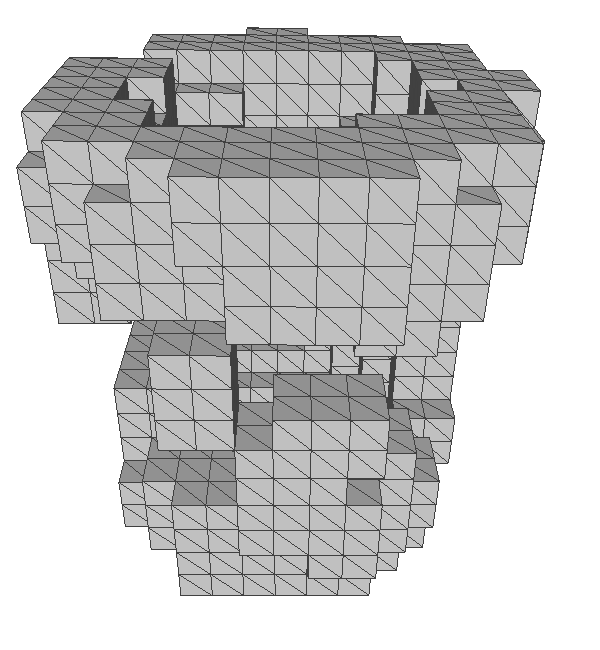} &
\includegraphics[width=.085\linewidth]{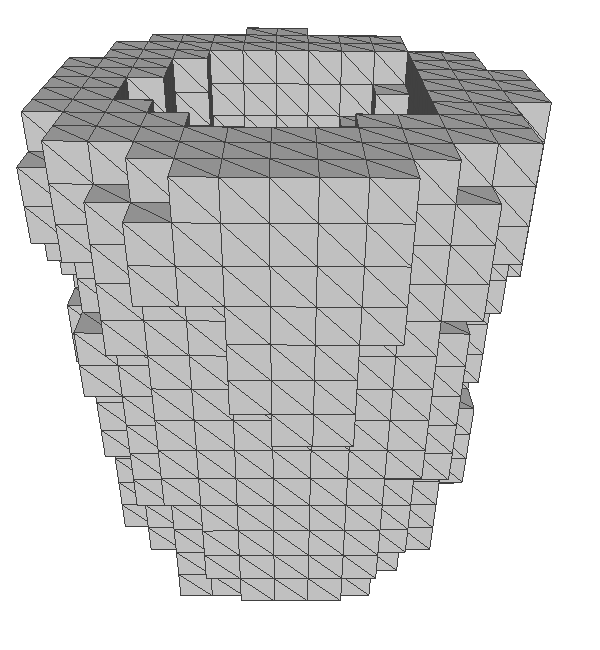} &
\includegraphics[width=.085\linewidth]{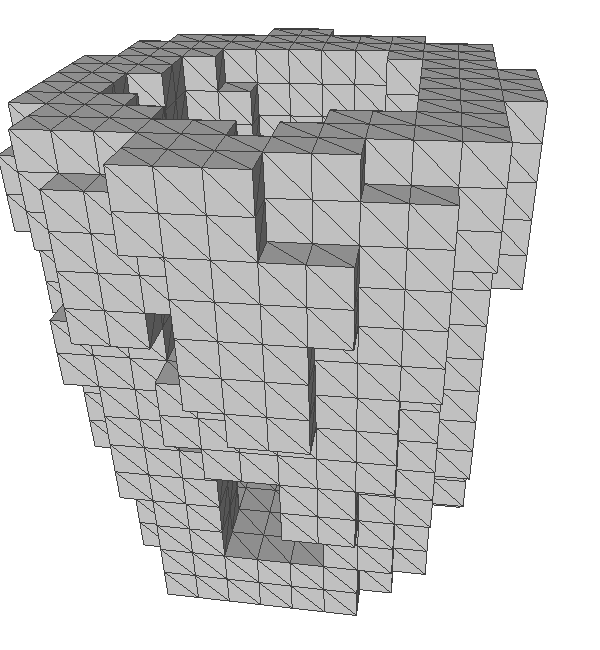} &
\includegraphics[width=.085\linewidth]{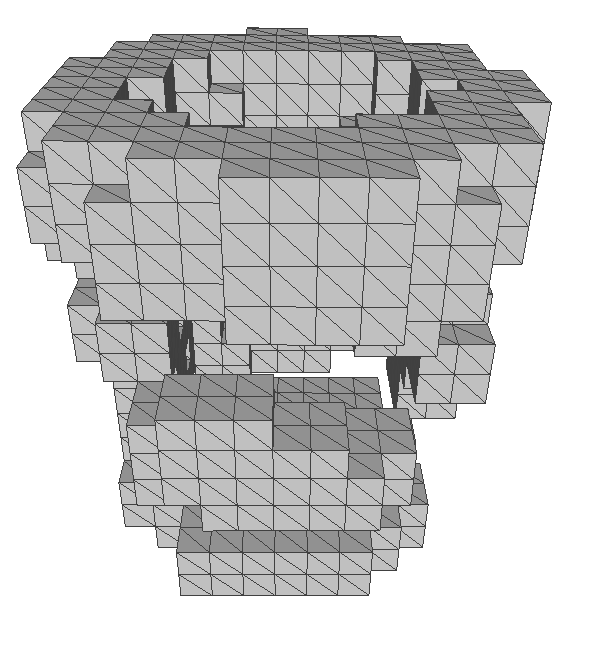} &
\includegraphics[width=.085\linewidth]{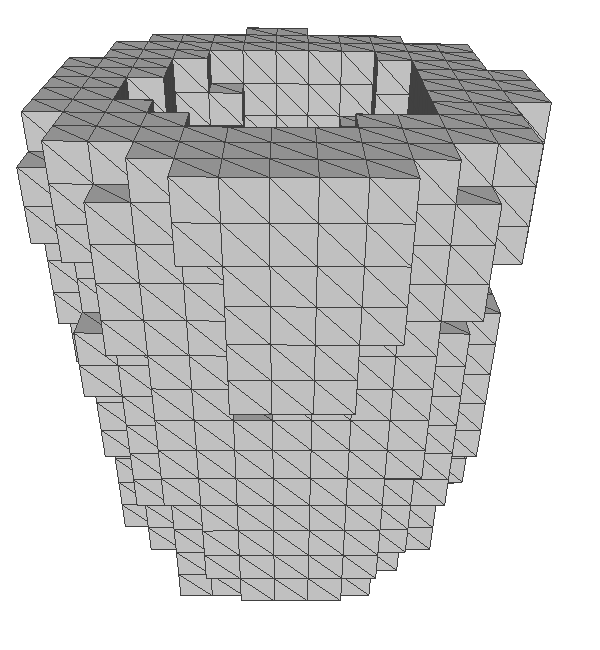} &
\includegraphics[width=.085\linewidth]{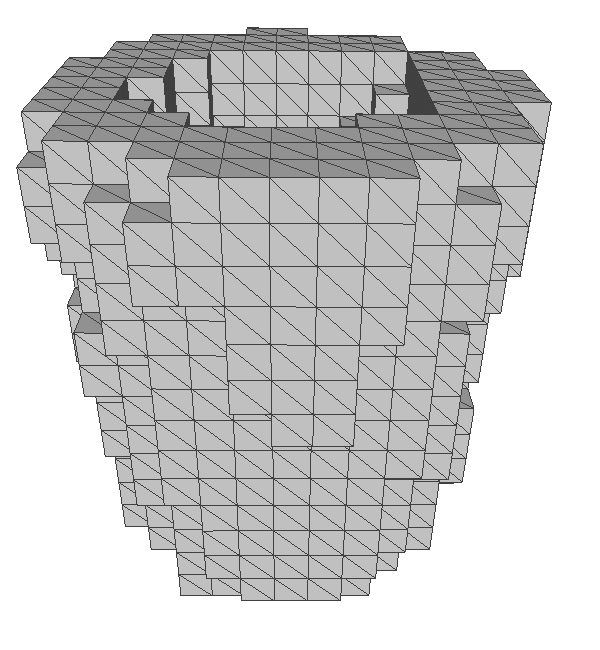} &
\includegraphics[width=.085\linewidth]{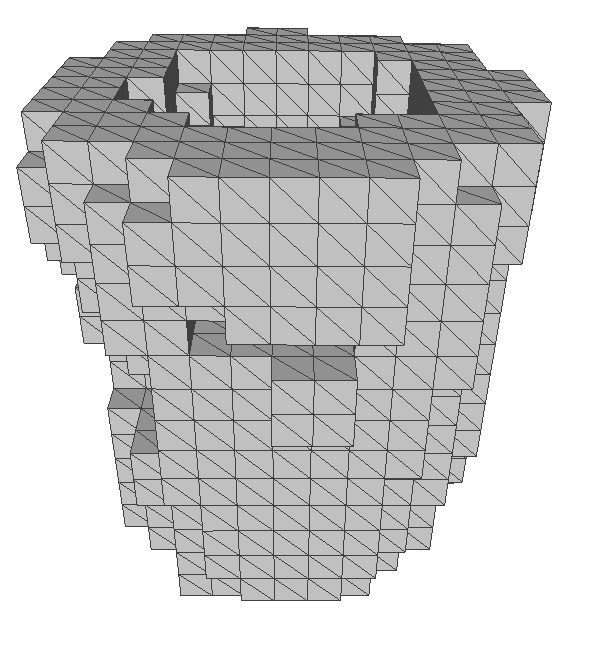} &
\includegraphics[width=.085\linewidth]{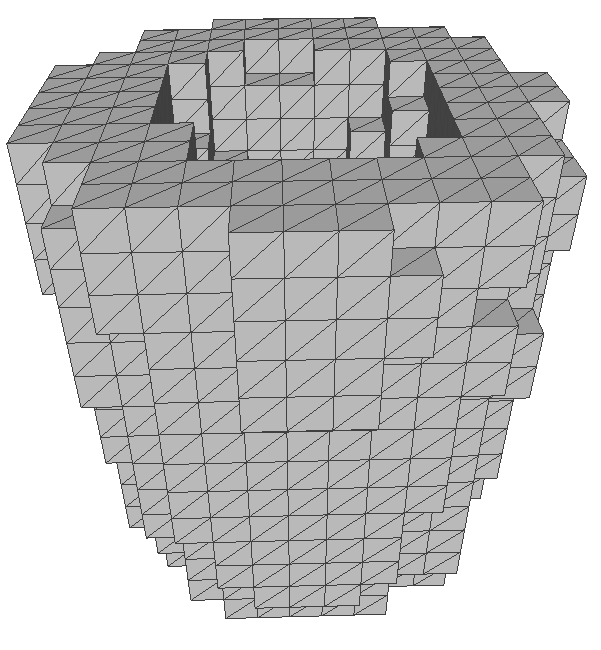} \\
{\begin{sideways} \centering \small{mesh} \end{sideways}} & 
\includegraphics[width=.085\linewidth]{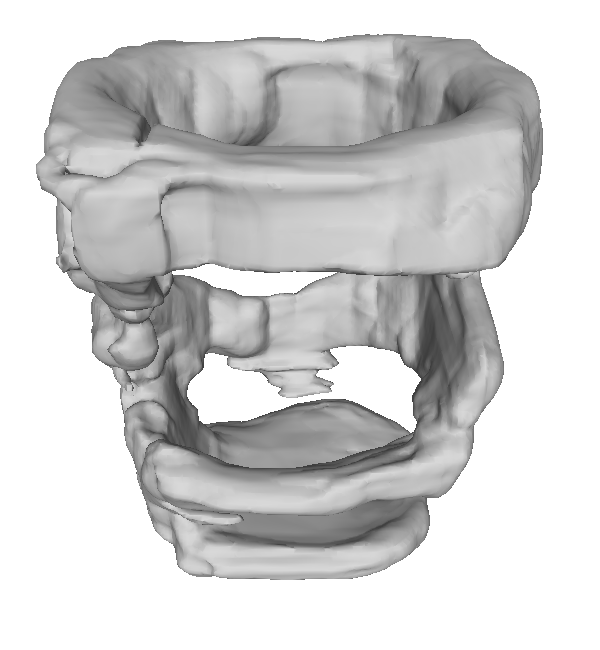} &
\includegraphics[width=.085\linewidth]{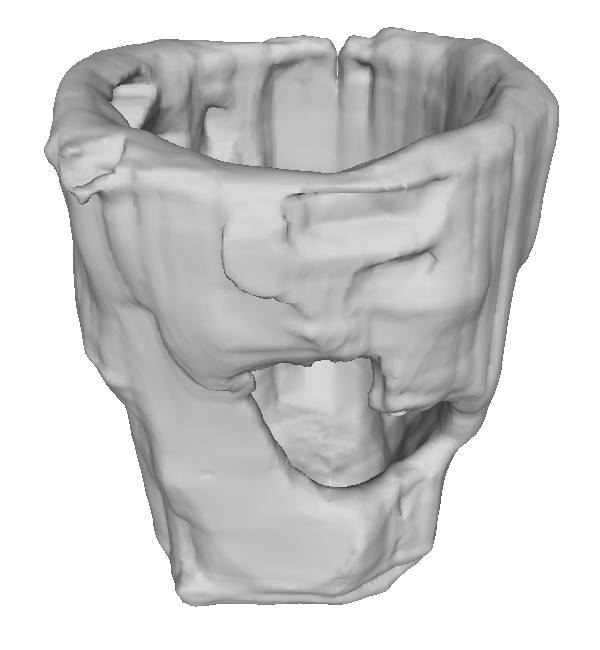} &
\includegraphics[width=.085\linewidth]{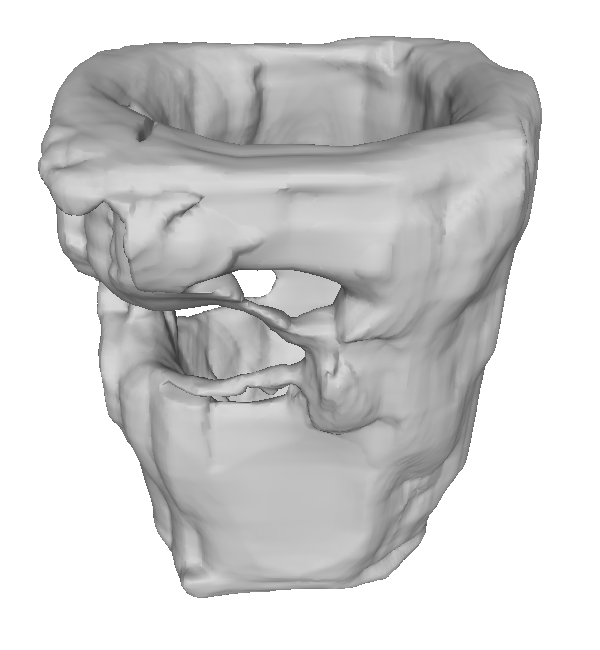} &
\includegraphics[width=.085\linewidth]{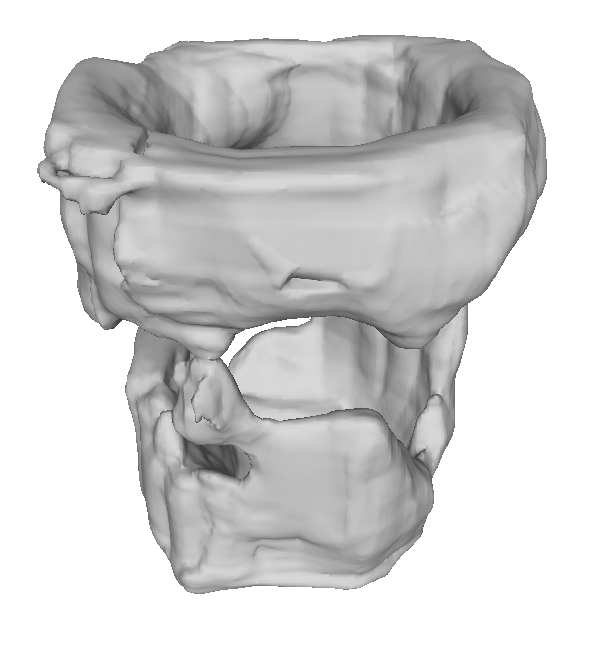} &
\includegraphics[width=.085\linewidth]{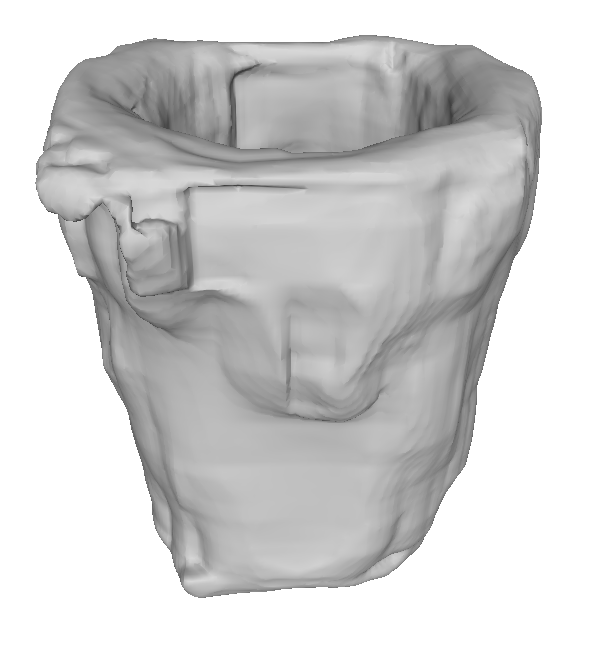} &
\includegraphics[width=.085\linewidth]{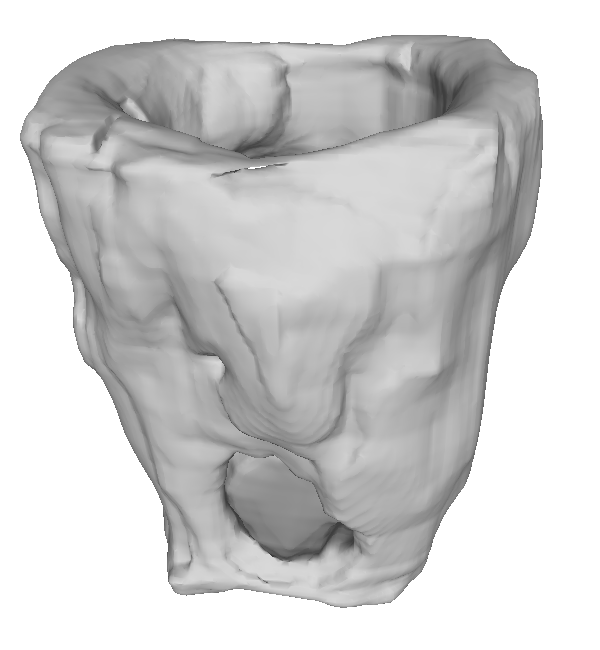} &
\includegraphics[width=.085\linewidth]{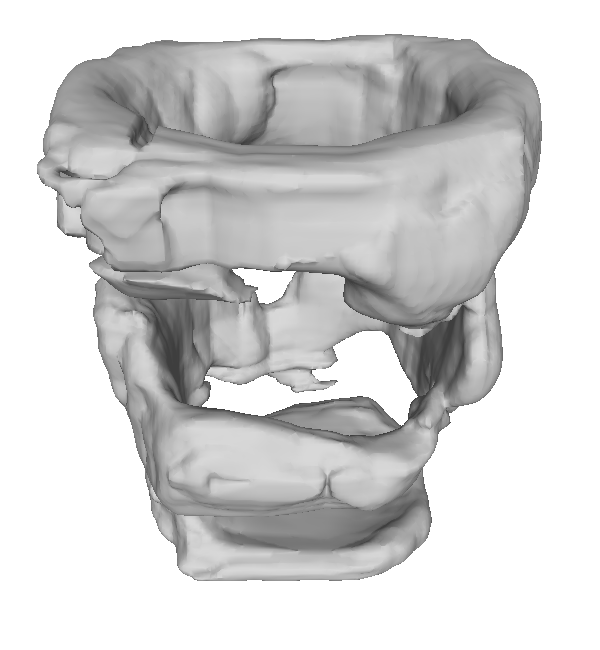} &
\includegraphics[width=.085\linewidth]{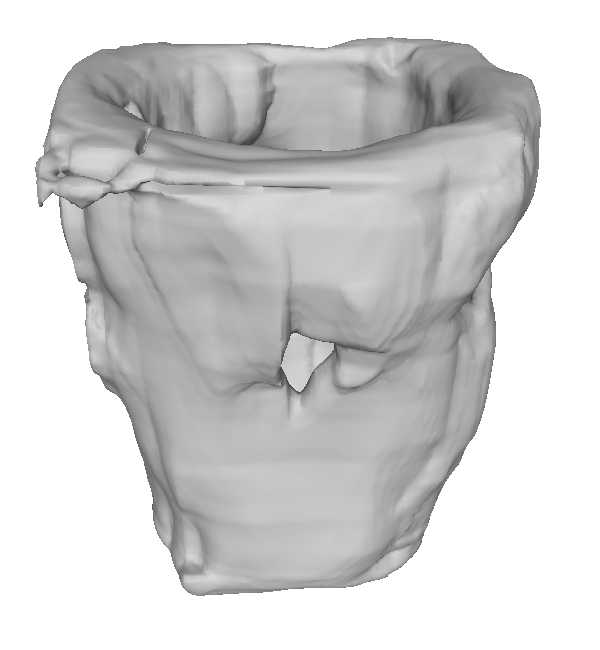} &
\includegraphics[width=.085\linewidth]{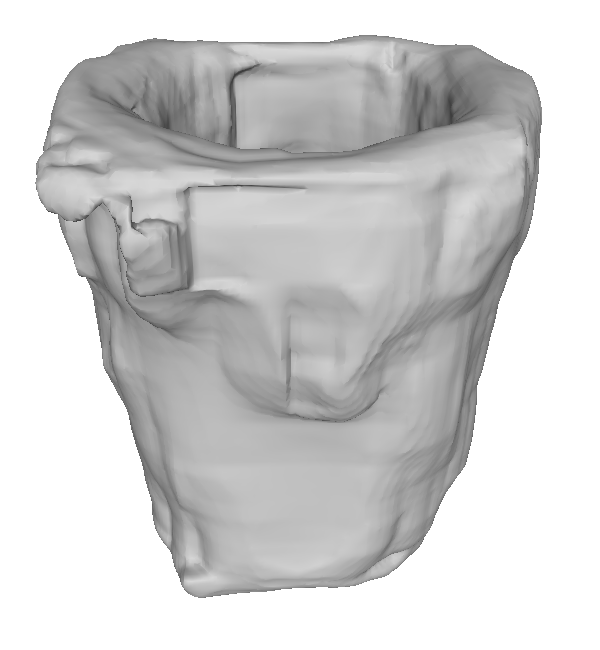} &
\includegraphics[width=.085\linewidth]{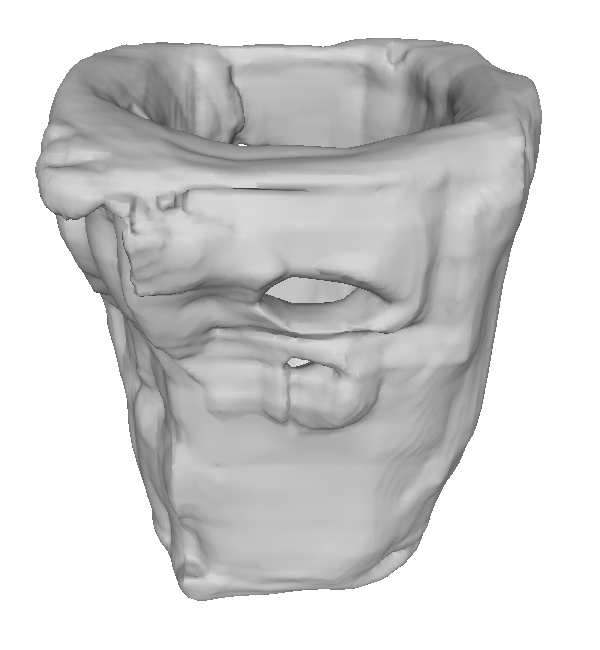} &
\includegraphics[width=.085\linewidth]{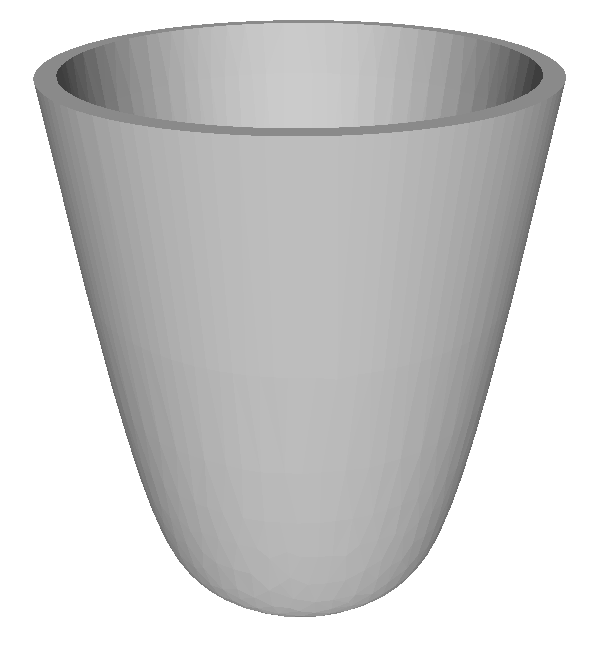} \\
\hline
{\begin{sideways} \centering \small{grid} \end{sideways}} &
\includegraphics[width=.085\linewidth]{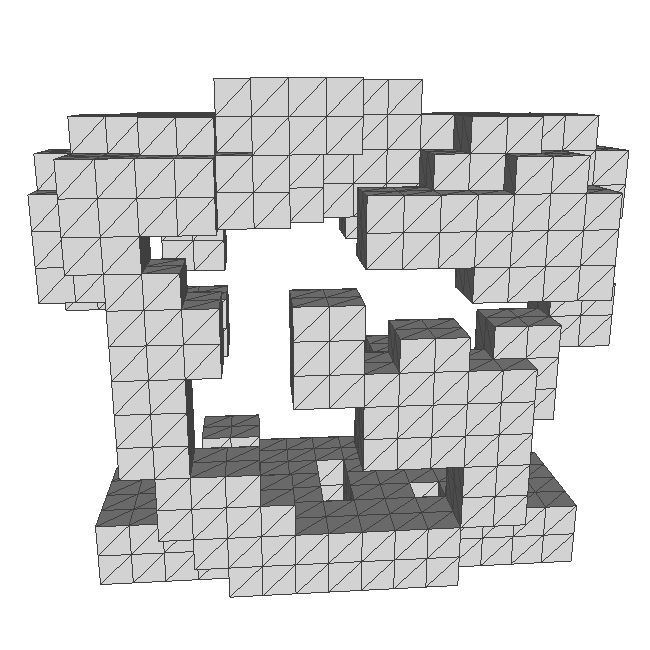} &
\includegraphics[width=.085\linewidth]{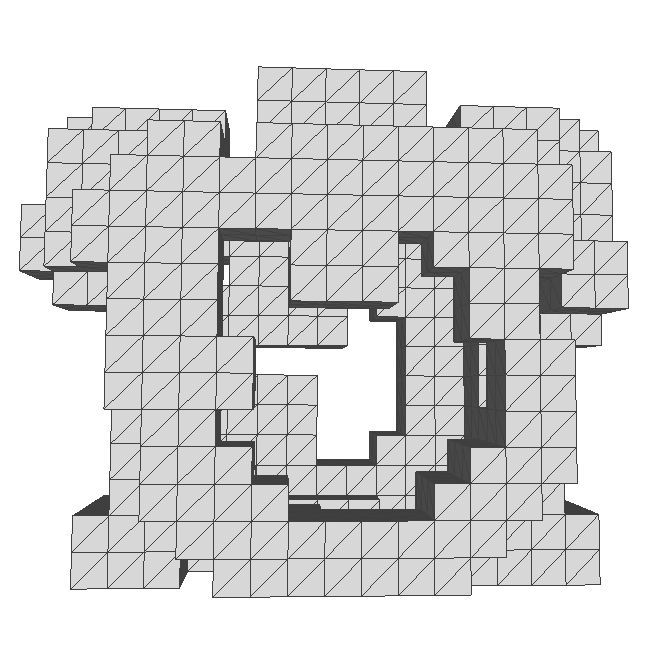} &
\includegraphics[width=.085\linewidth]{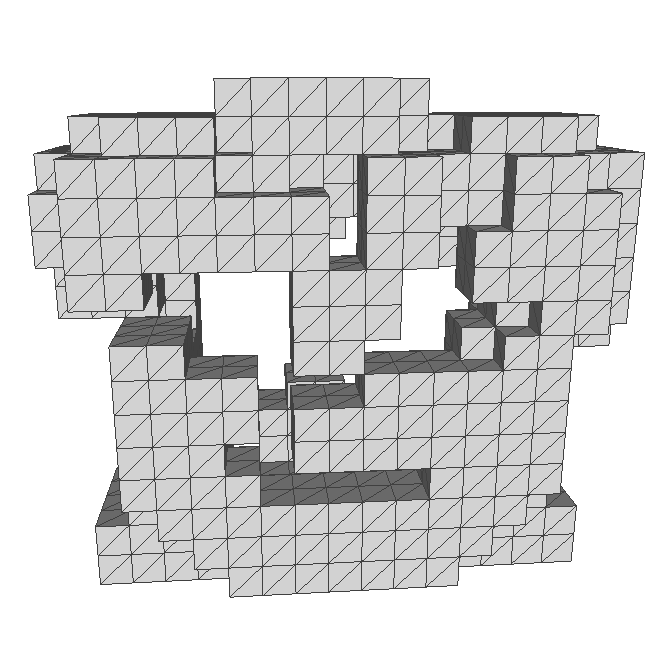} &
\includegraphics[width=.085\linewidth]{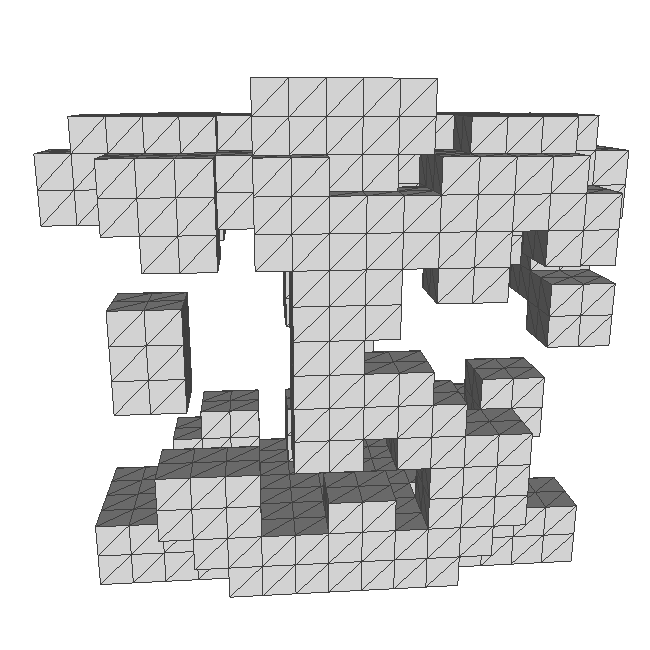} &
\includegraphics[width=.085\linewidth]{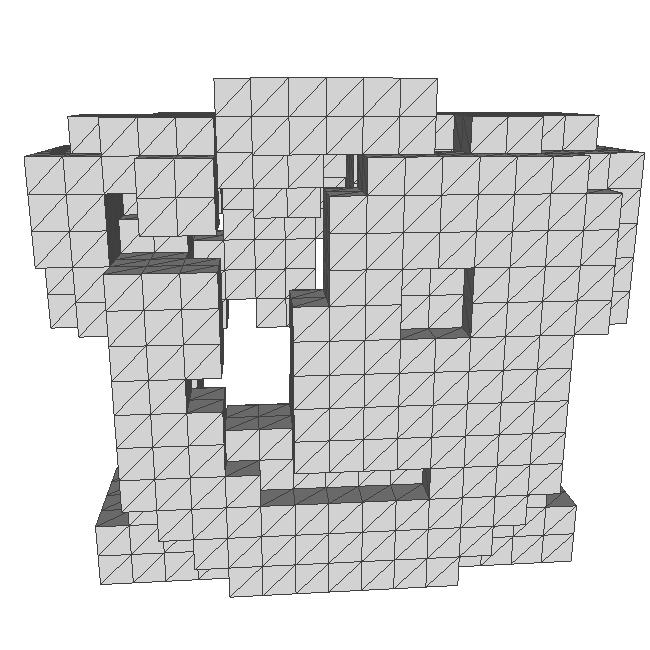} &
\includegraphics[width=.085\linewidth]{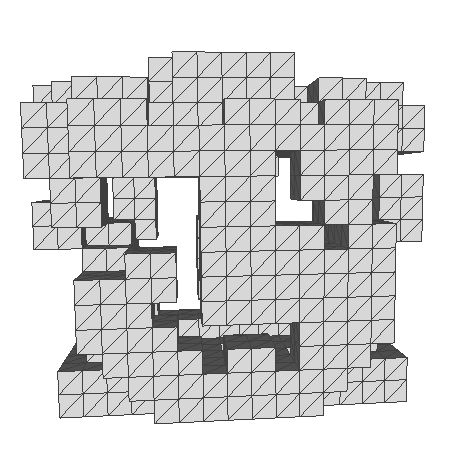} &
\includegraphics[width=.085\linewidth]{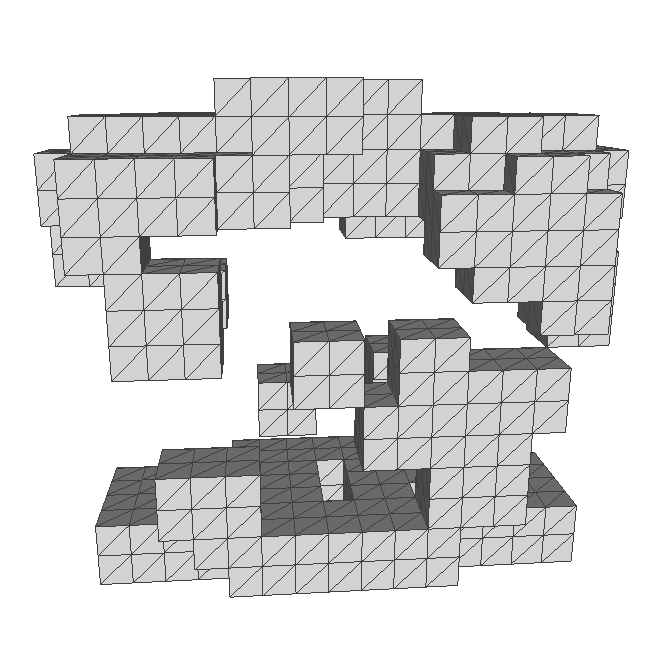} &
\includegraphics[width=.085\linewidth]{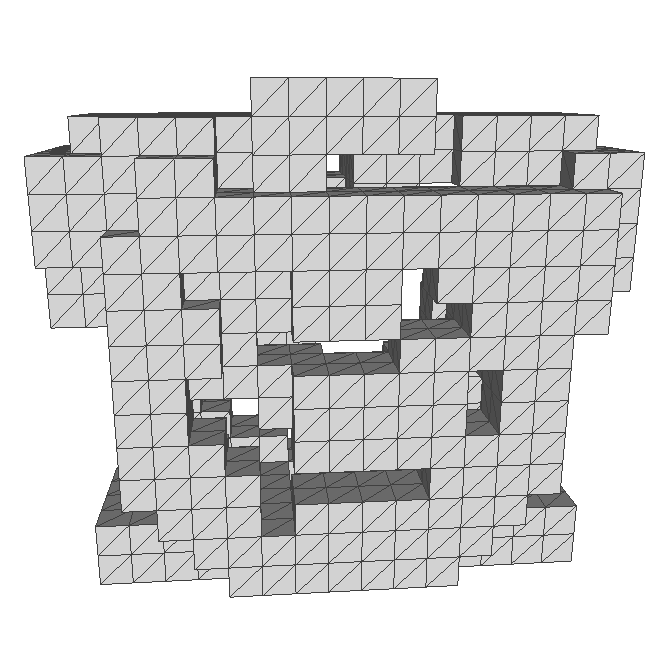} &
\includegraphics[width=.085\linewidth]{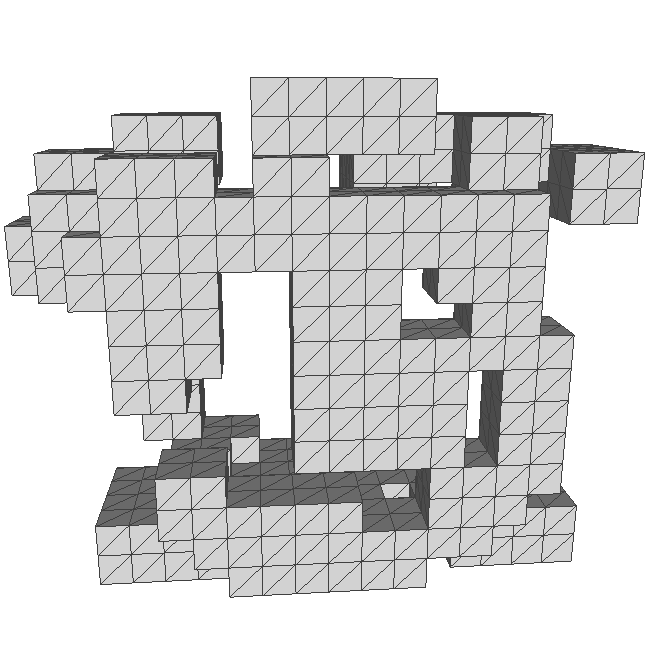} &
\includegraphics[width=.085\linewidth]{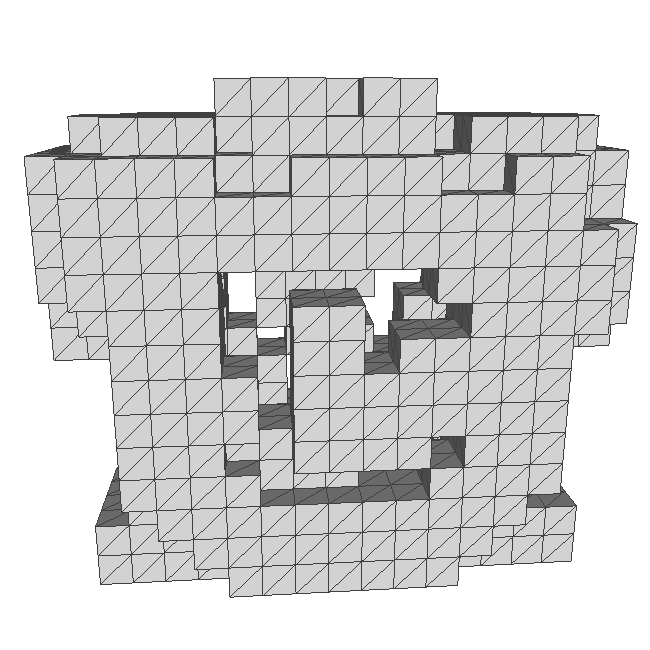} &
\includegraphics[width=.085\linewidth]{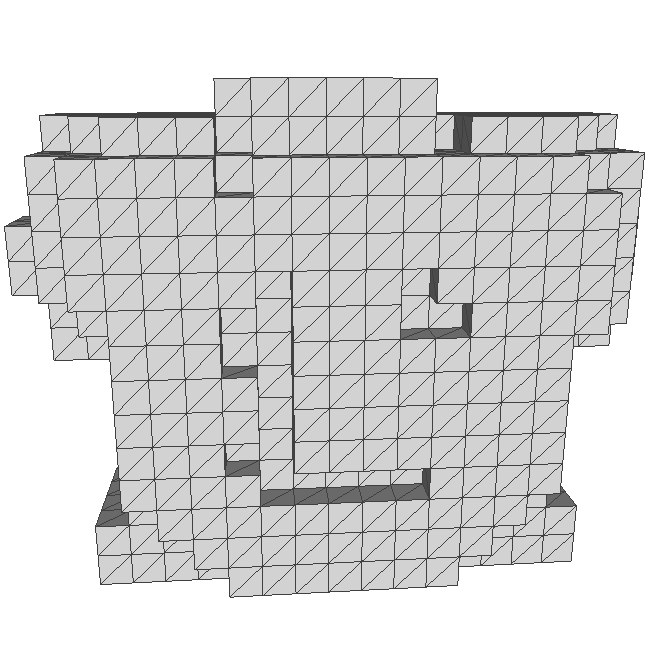} \\
{\begin{sideways} \centering \small{mesh} \end{sideways}} &
\includegraphics[width=.085\linewidth]{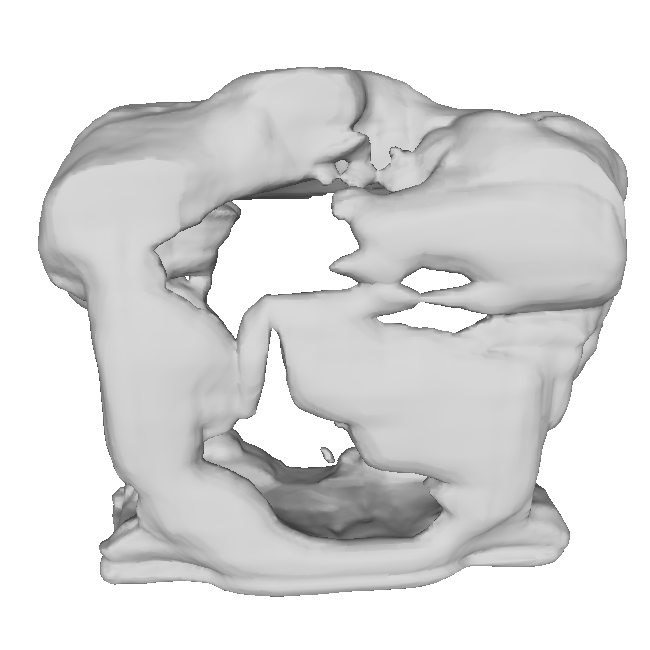} &
\includegraphics[width=.085\linewidth]{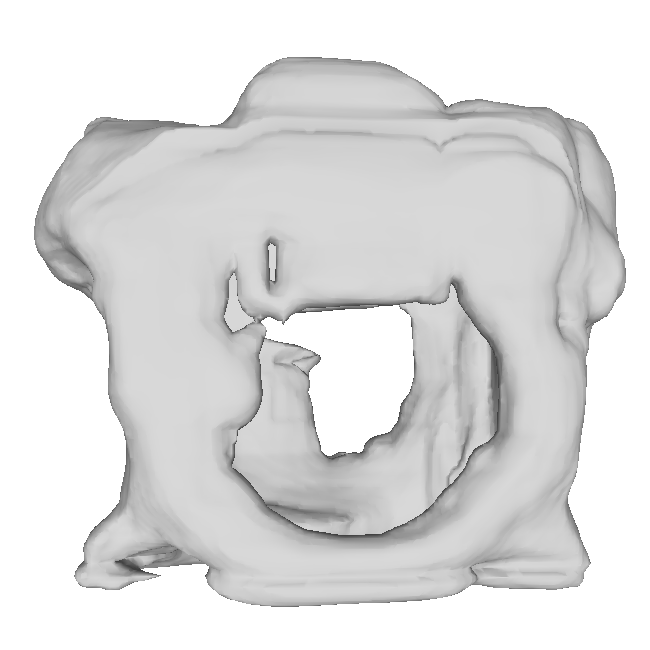} &
\includegraphics[width=.085\linewidth]{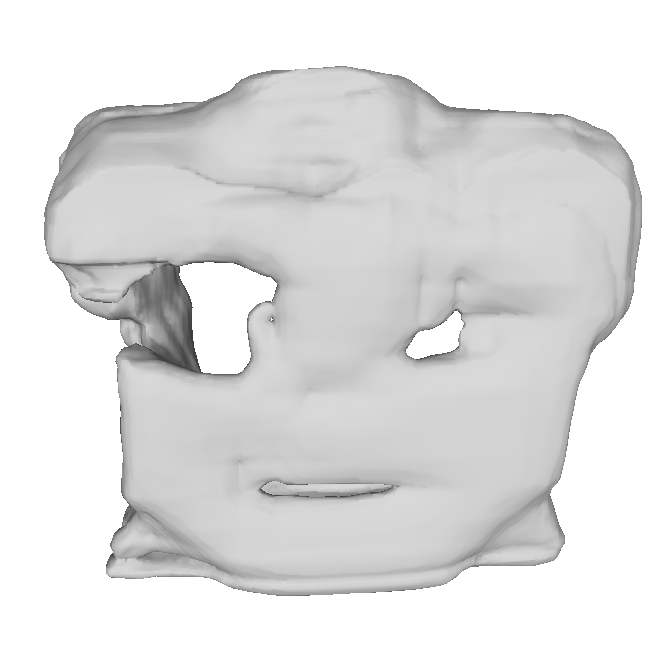} &
\includegraphics[width=.085\linewidth]{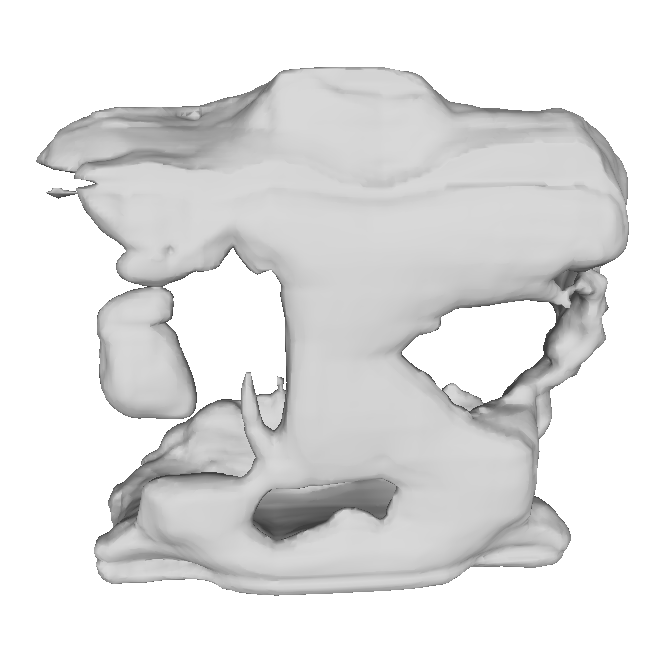} &
\includegraphics[width=.085\linewidth]{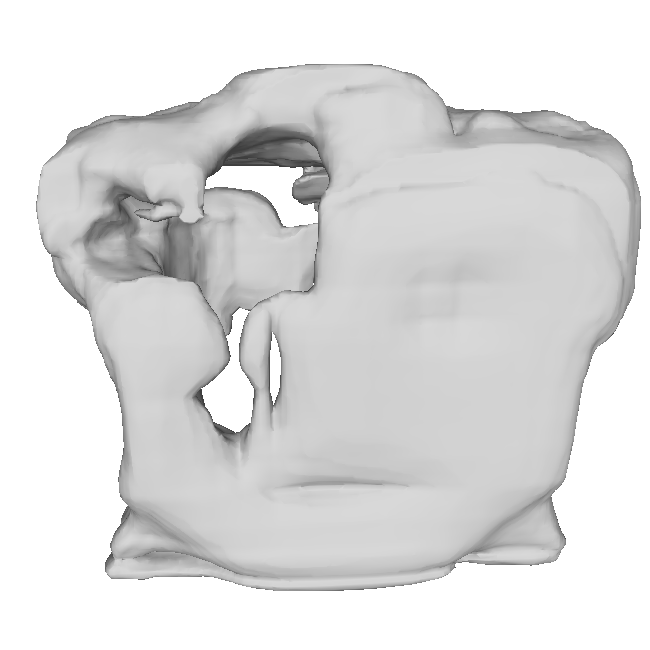} &
\includegraphics[width=.085\linewidth]{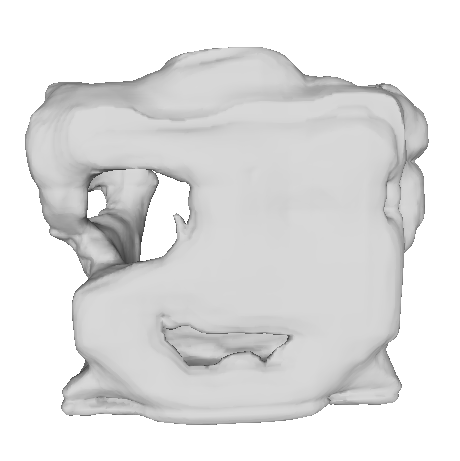} &
\includegraphics[width=.085\linewidth]{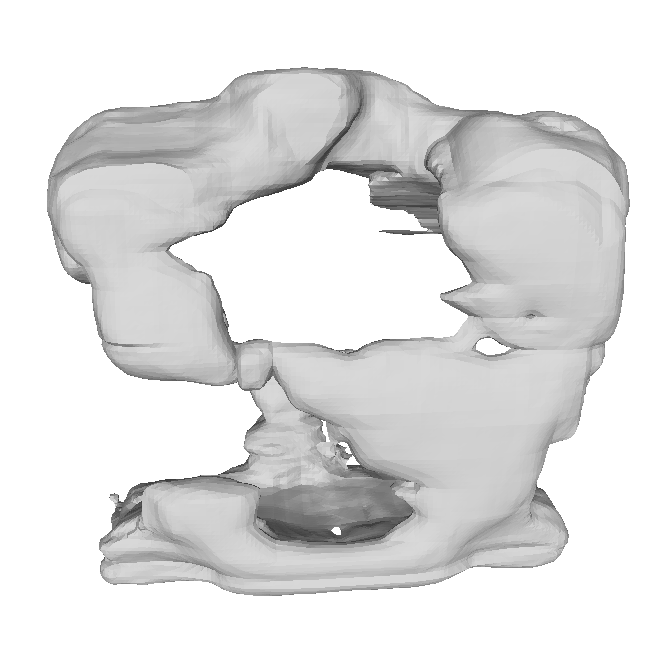} &
\includegraphics[width=.085\linewidth]{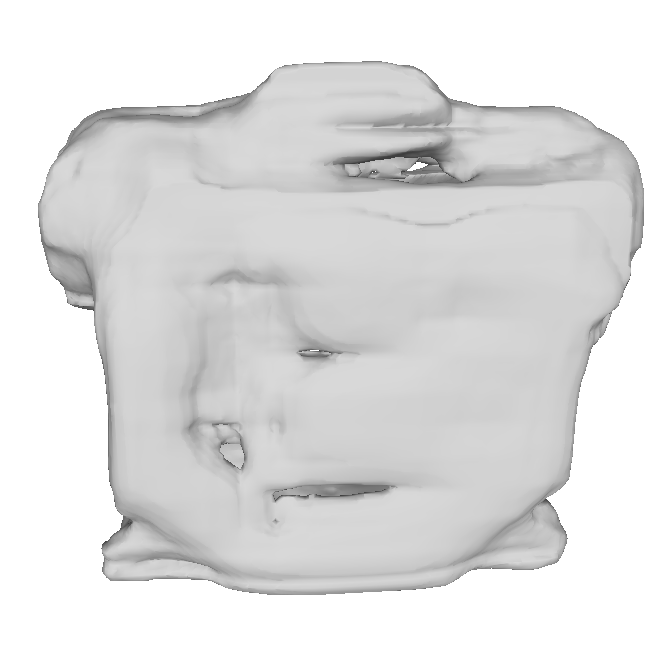} &
\includegraphics[width=.085\linewidth]{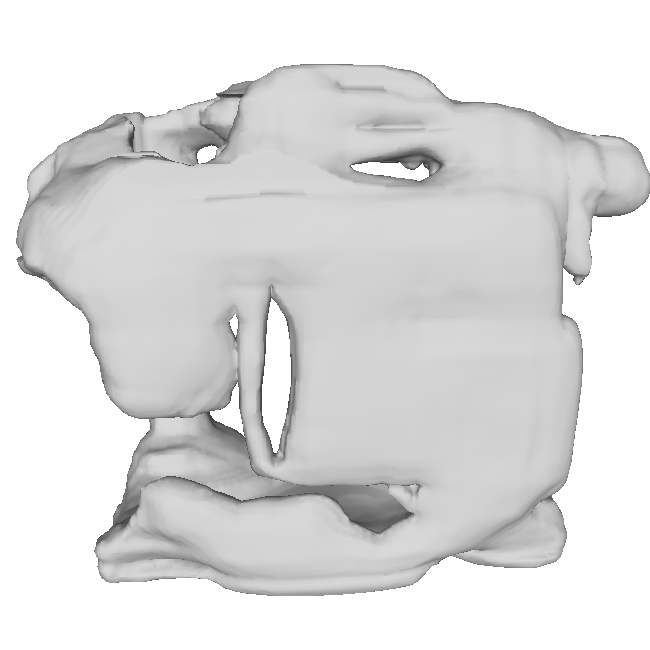} &
\includegraphics[width=.085\linewidth]{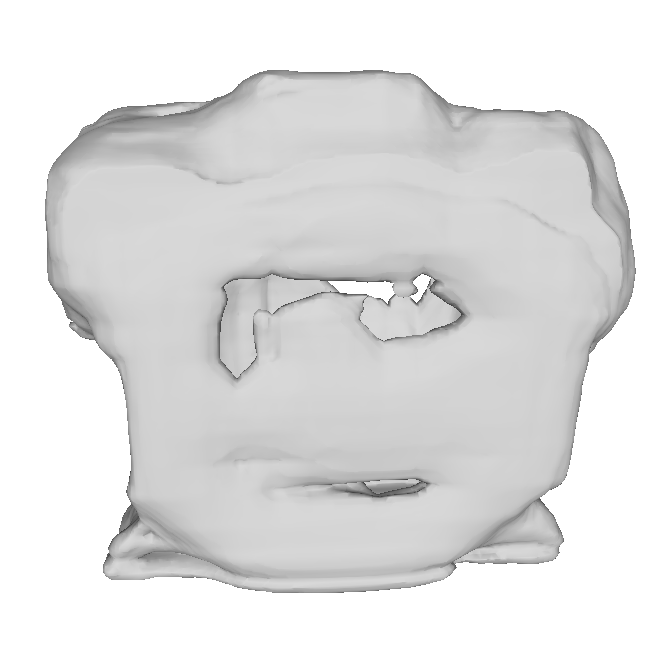} &
\includegraphics[width=.085\linewidth]{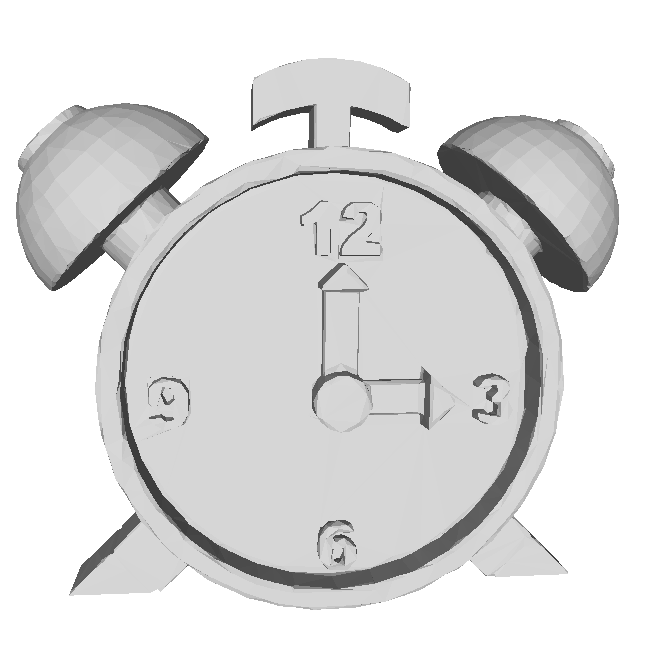} \\
\hline
{\begin{sideways} \centering \small{grid} \end{sideways}} &
\includegraphics[width=.085\linewidth]{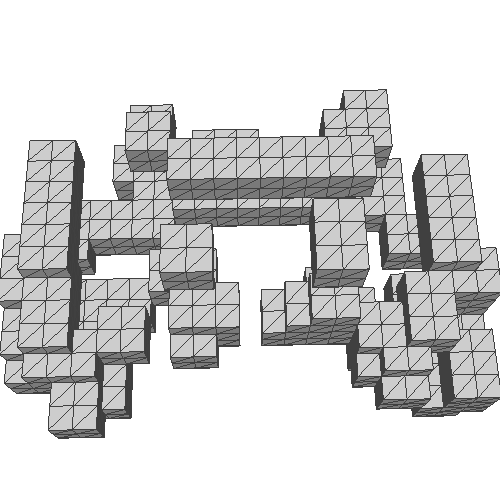} &
\includegraphics[width=.085\linewidth]{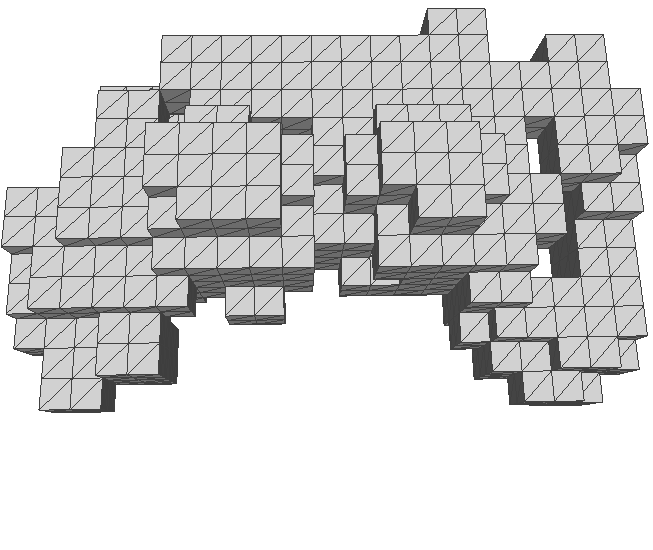} &
\includegraphics[width=.085\linewidth]{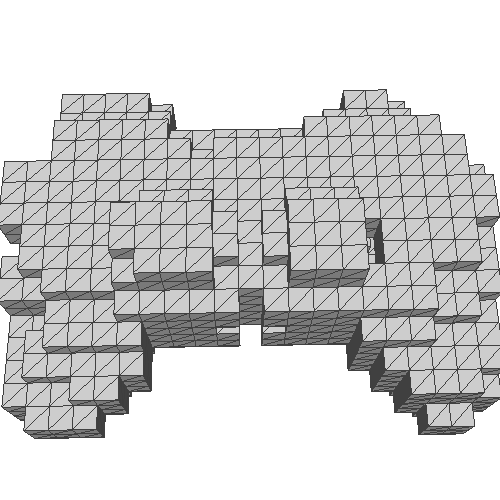} &
\includegraphics[width=.085\linewidth]{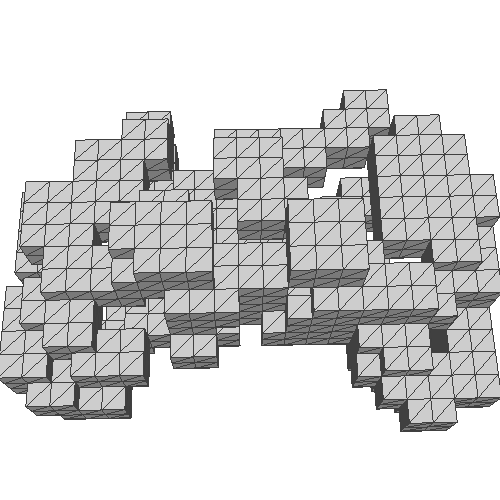} &
\includegraphics[width=.085\linewidth]{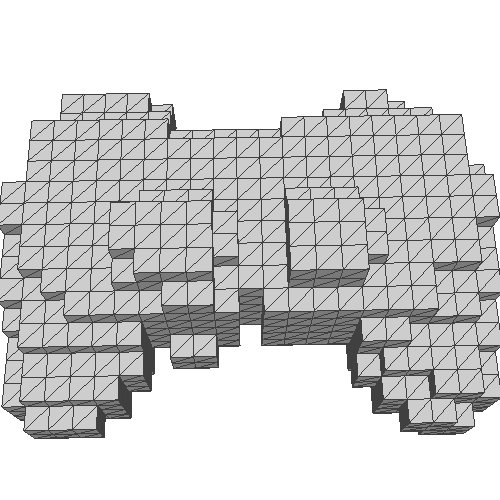} &
\includegraphics[width=.085\linewidth]{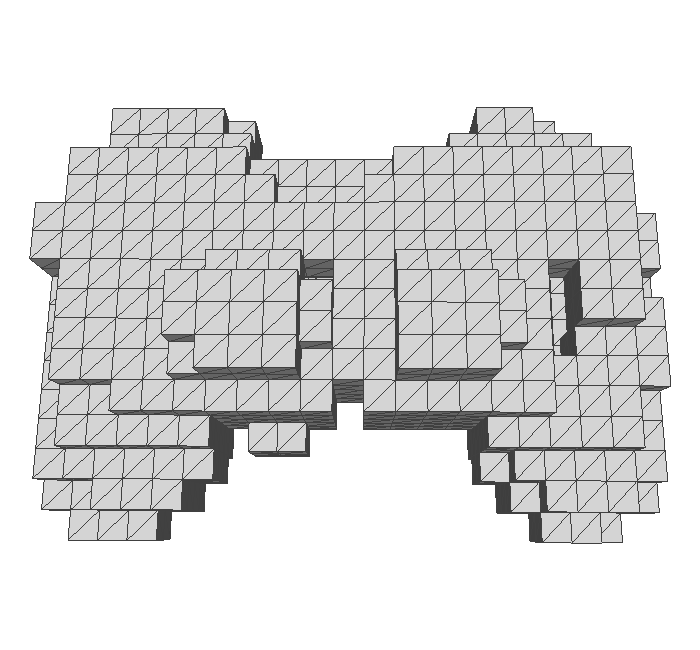} &
\includegraphics[width=.085\linewidth]{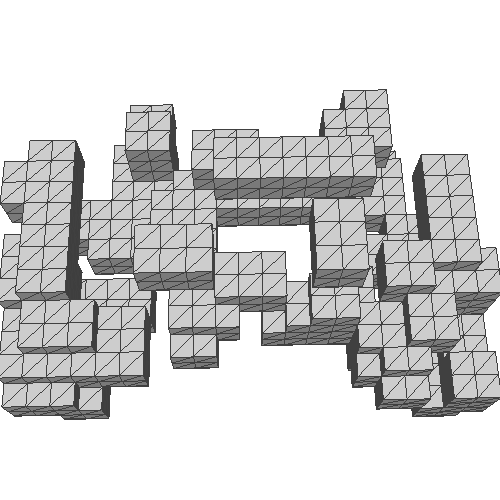} &
\includegraphics[width=.085\linewidth]{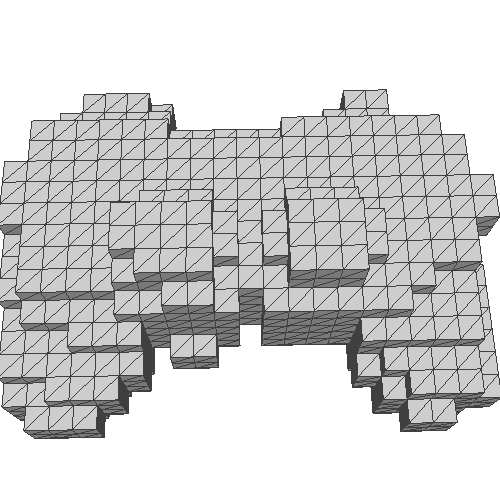} &
\includegraphics[width=.085\linewidth]{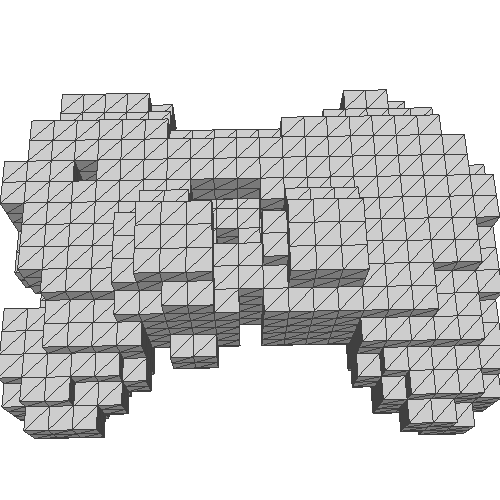} &
\includegraphics[width=.085\linewidth]{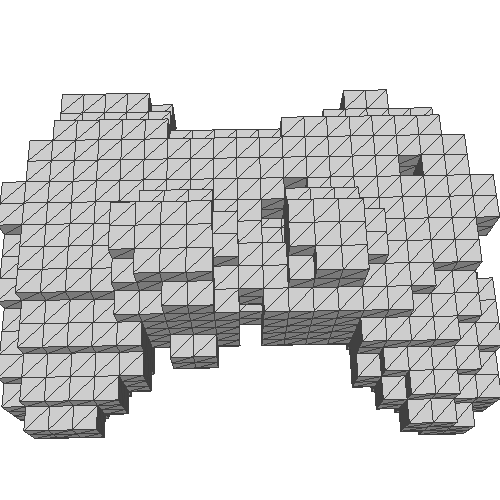} &
\includegraphics[width=.085\linewidth]{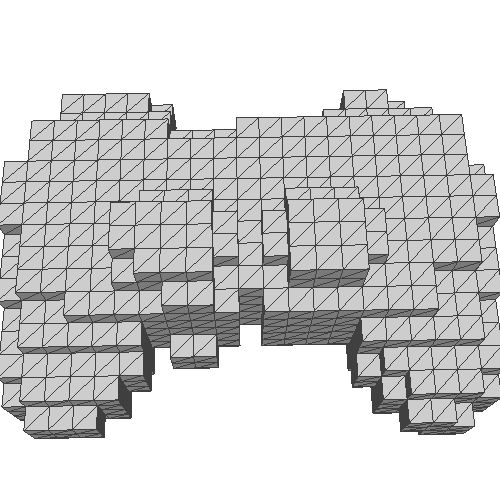} \\
{\begin{sideways} \centering \small{mesh} \end{sideways}} &
\includegraphics[width=.085\linewidth]{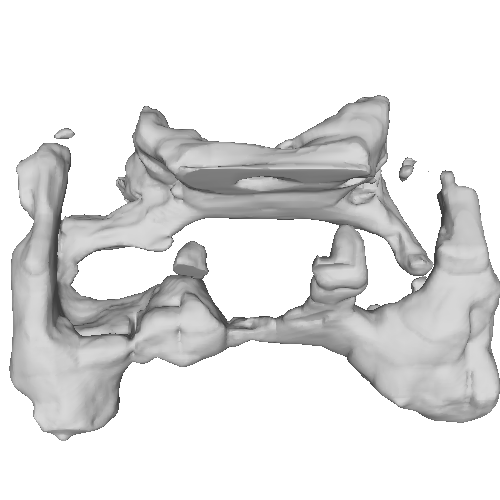} &
\includegraphics[width=.085\linewidth]{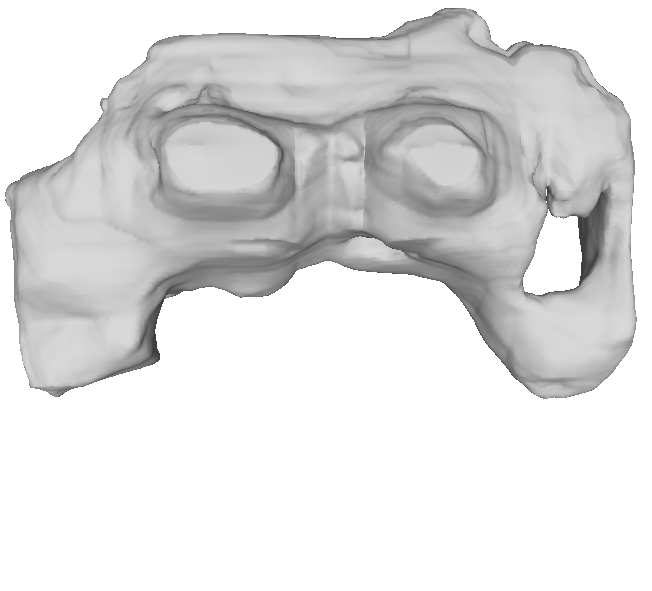} &
\includegraphics[width=.085\linewidth]{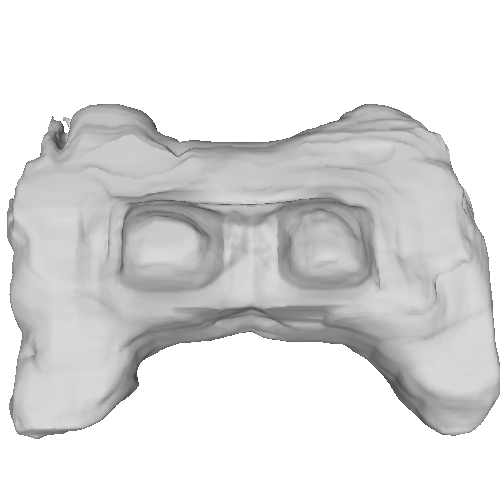} &
\includegraphics[width=.085\linewidth]{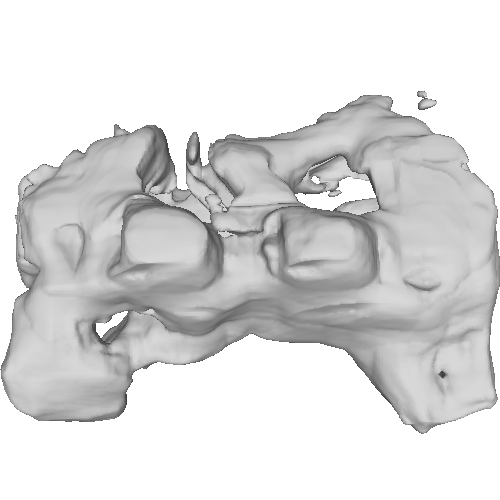} &
\includegraphics[width=.085\linewidth]{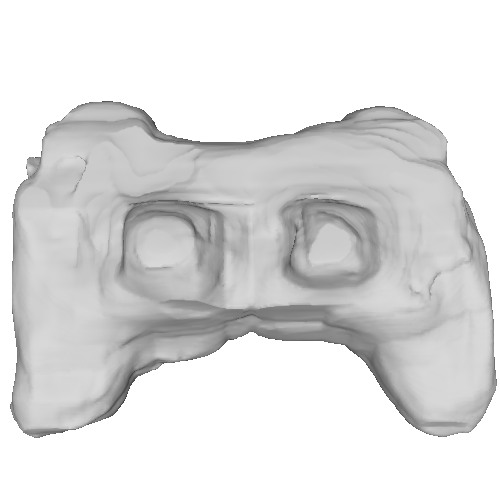} &
\includegraphics[width=.085\linewidth]{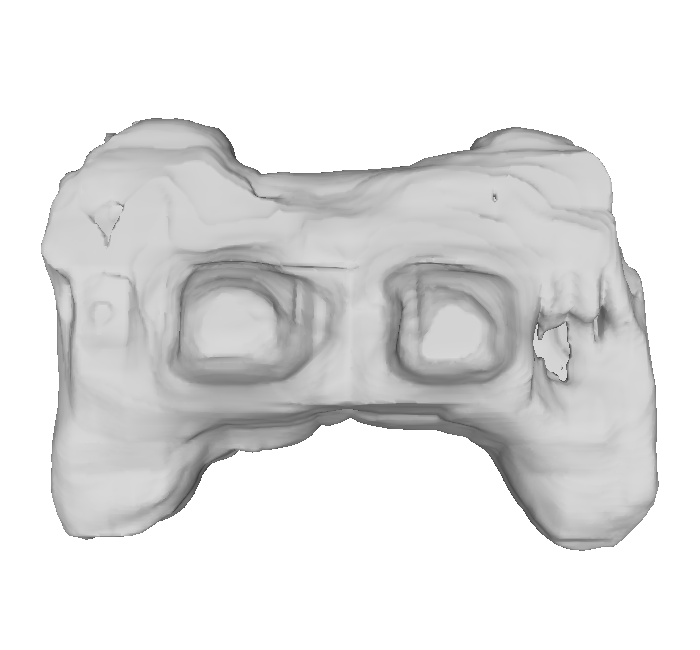} &
\includegraphics[width=.085\linewidth]{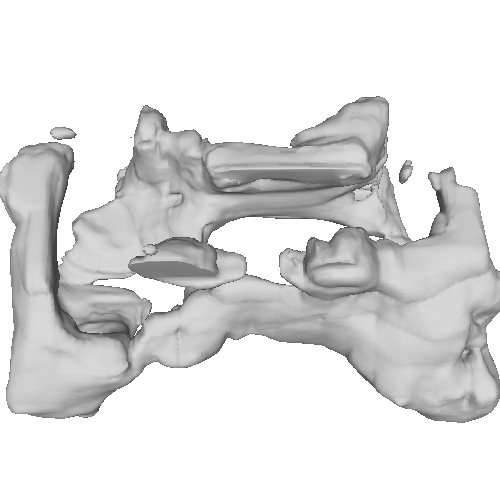} &
\includegraphics[width=.085\linewidth]{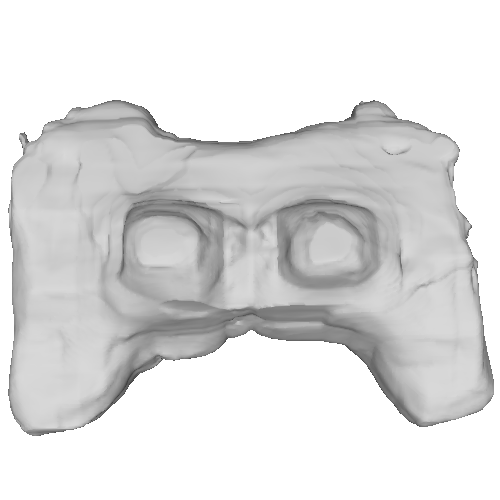} &
\includegraphics[width=.085\linewidth]{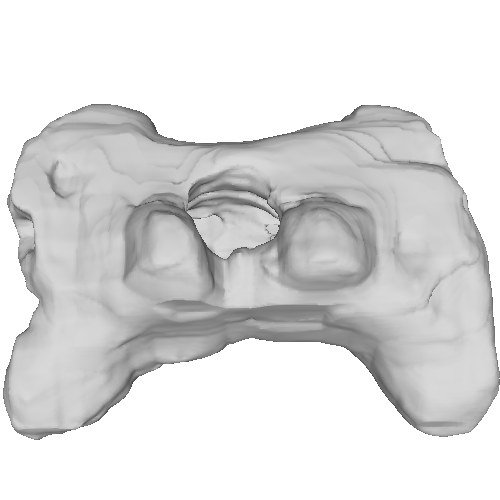} &
\includegraphics[width=.085\linewidth]{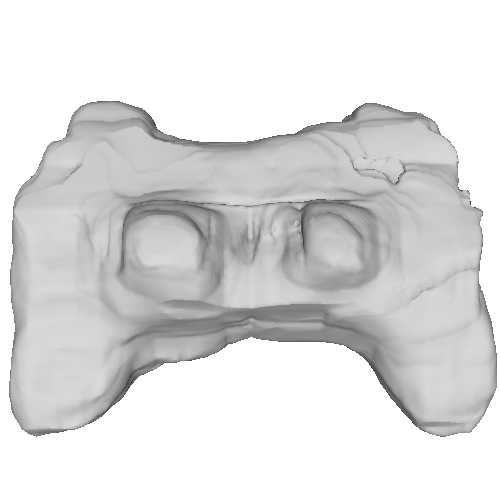} &
\includegraphics[width=.085\linewidth]{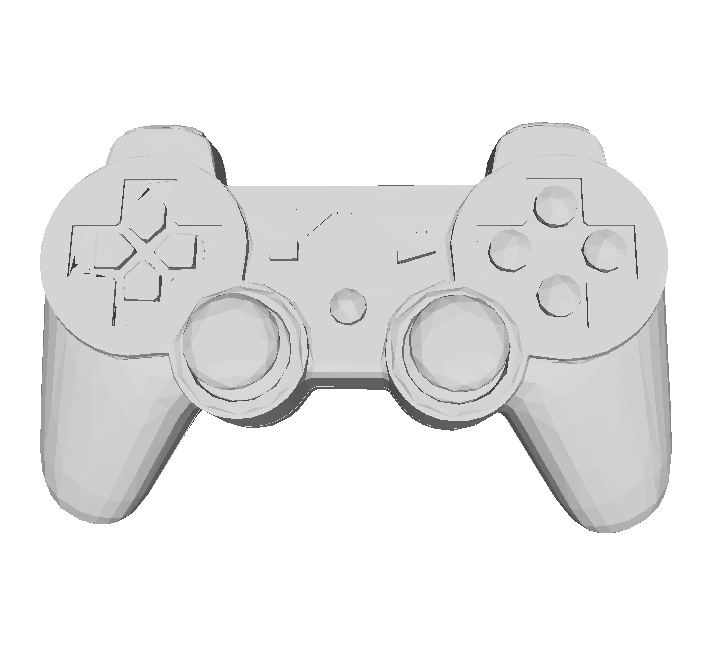} \\
\hline
{\begin{sideways} \centering \small{grid} \end{sideways}} &
\includegraphics[width=.085\linewidth]{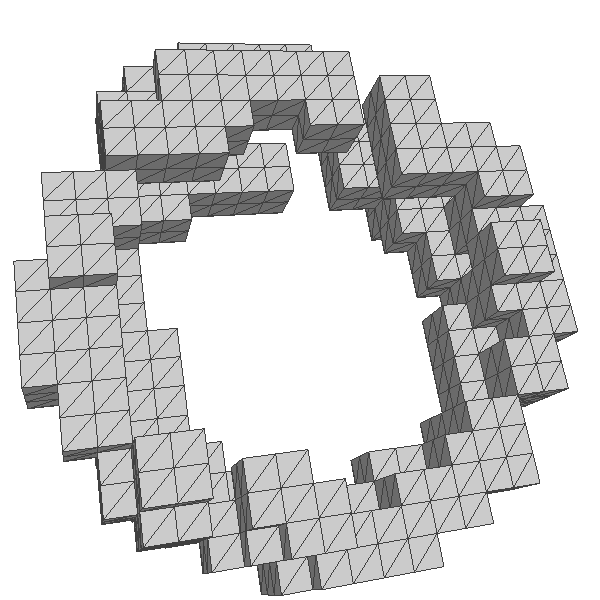} &
\includegraphics[width=.085\linewidth]{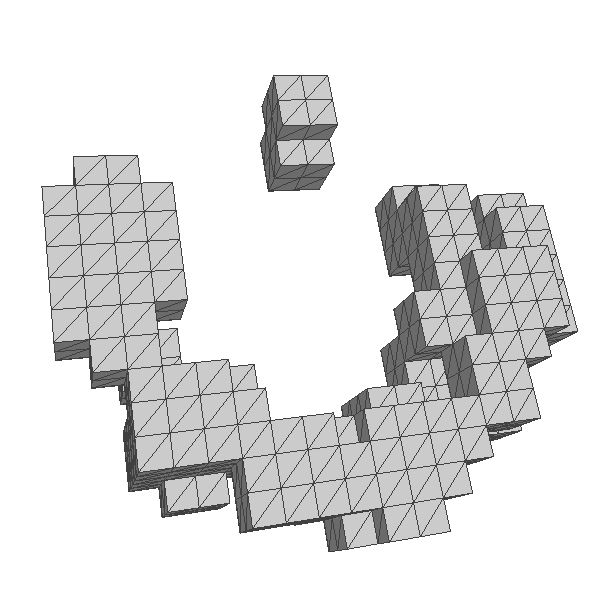} &
\includegraphics[width=.085\linewidth]{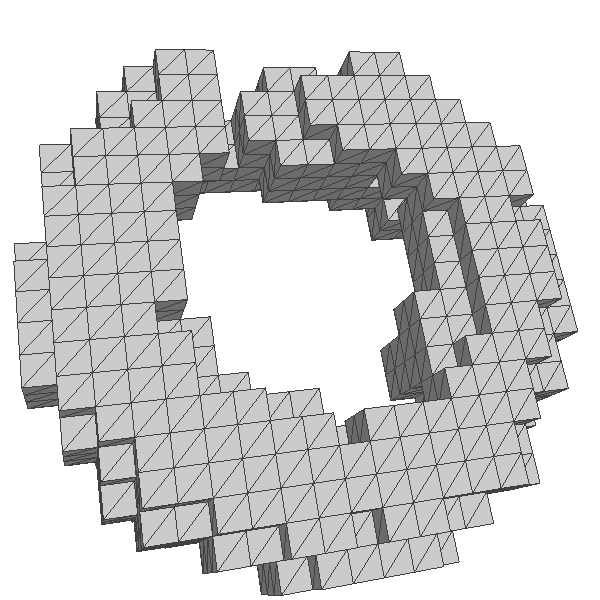} &
\includegraphics[width=.085\linewidth]{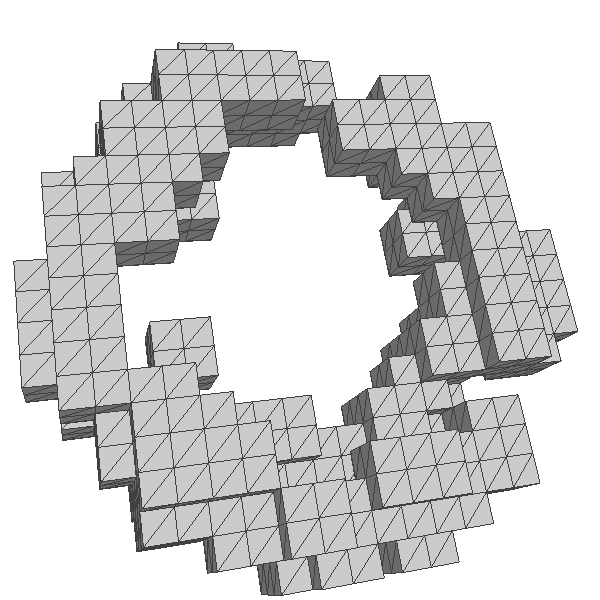} &
\includegraphics[width=.085\linewidth]{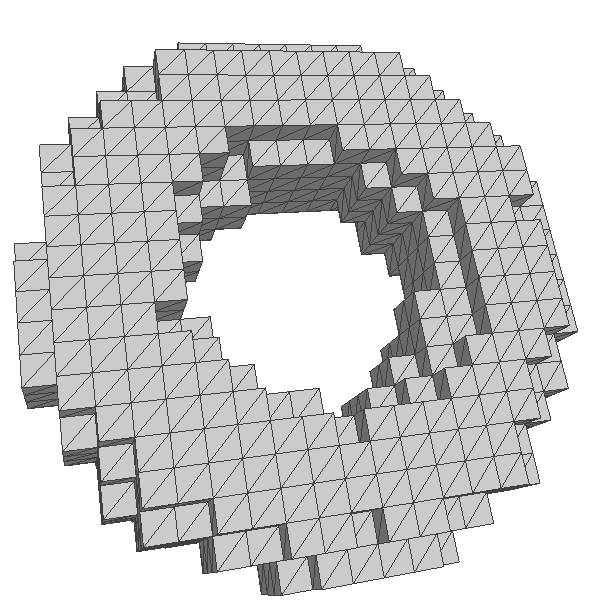} &
\includegraphics[width=.085\linewidth]{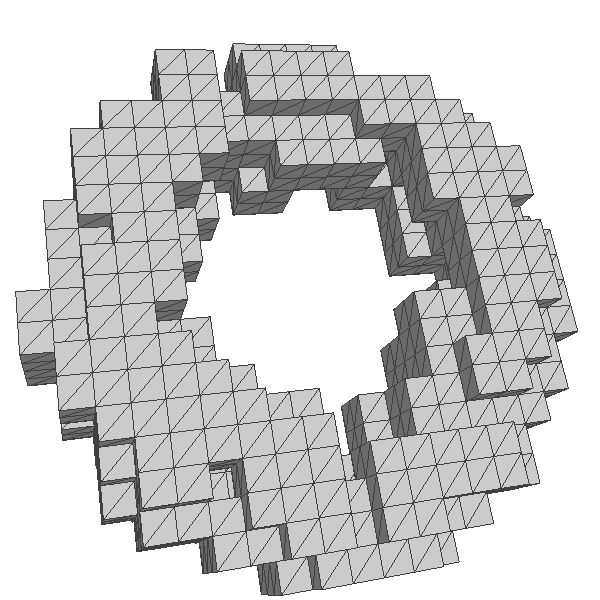} &
\includegraphics[width=.085\linewidth]{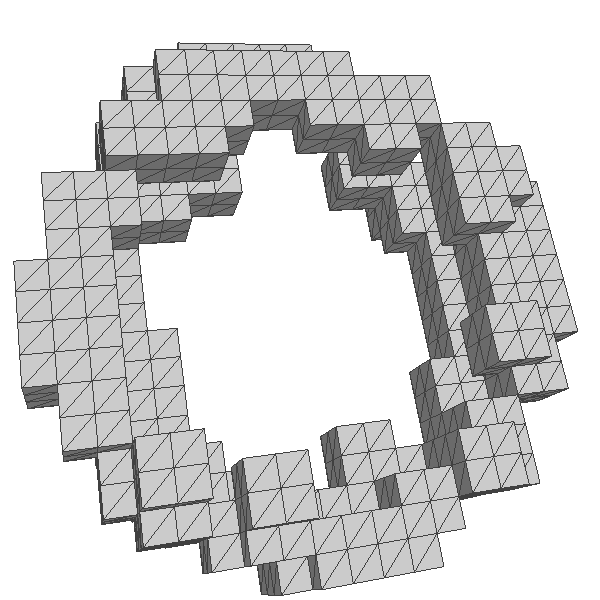} &
\includegraphics[width=.085\linewidth]{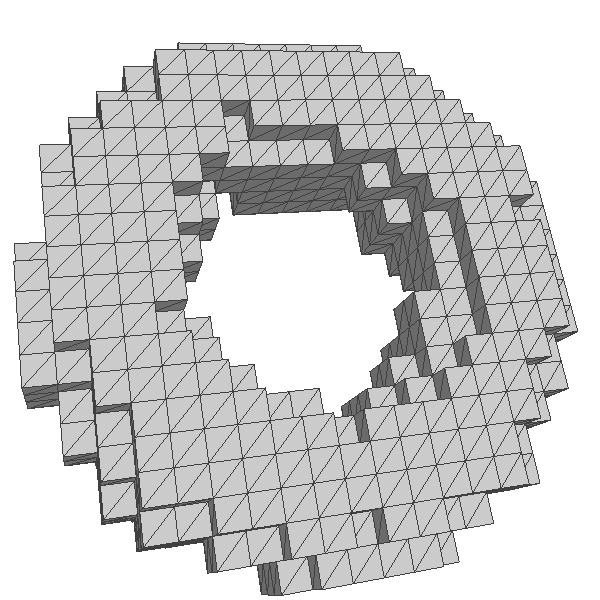} &
\includegraphics[width=.085\linewidth]{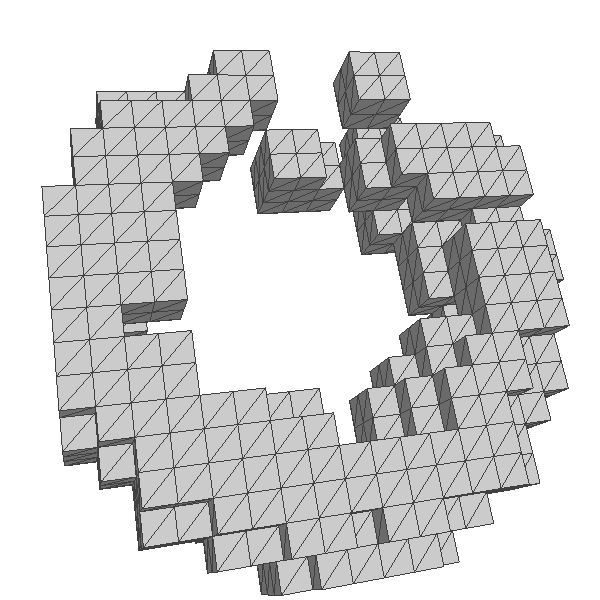} &
\includegraphics[width=.085\linewidth]{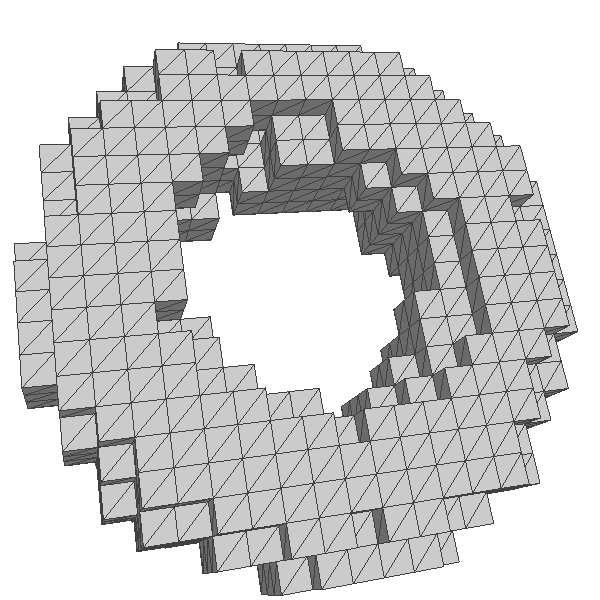} &
\includegraphics[width=.085\linewidth]{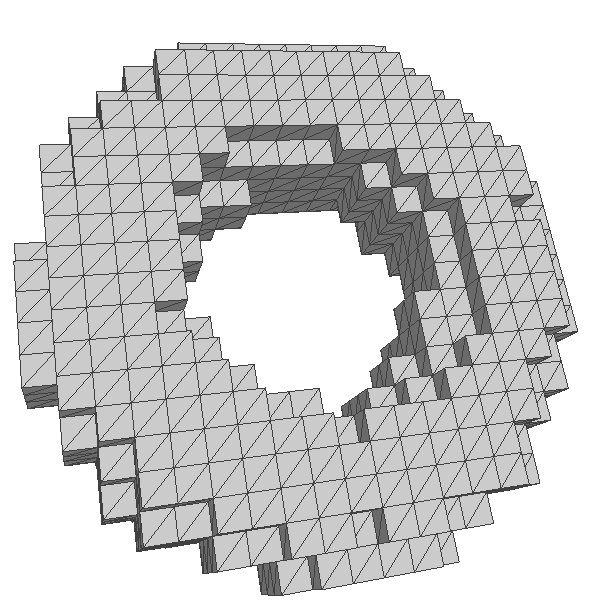} \\
{\begin{sideways} \centering \small{mesh} \end{sideways}} &
\includegraphics[width=.085\linewidth]{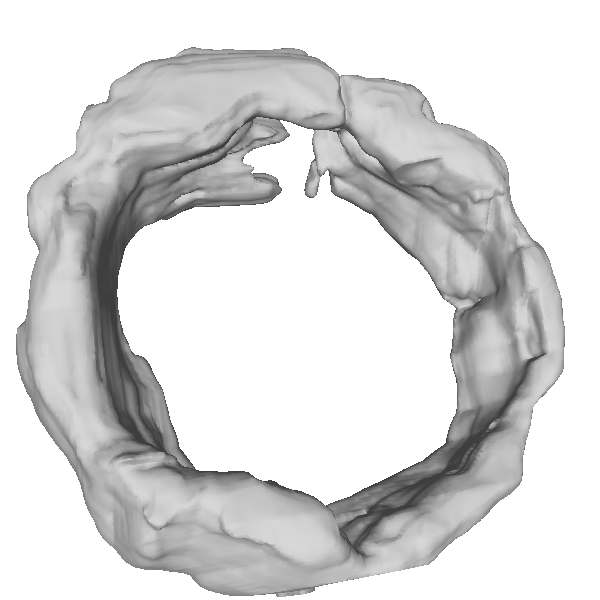} &
\includegraphics[width=.085\linewidth]{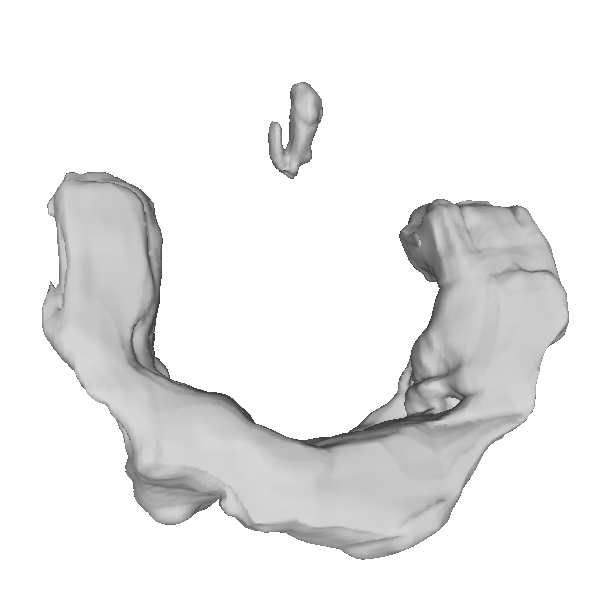} &
\includegraphics[width=.085\linewidth]{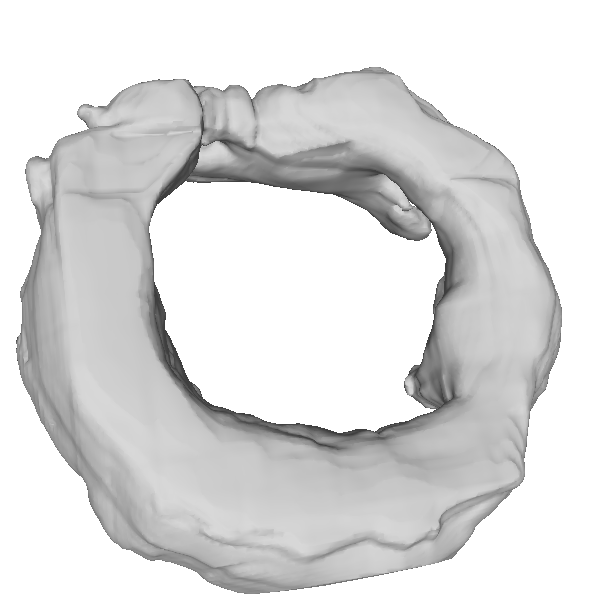} &
\includegraphics[width=.085\linewidth]{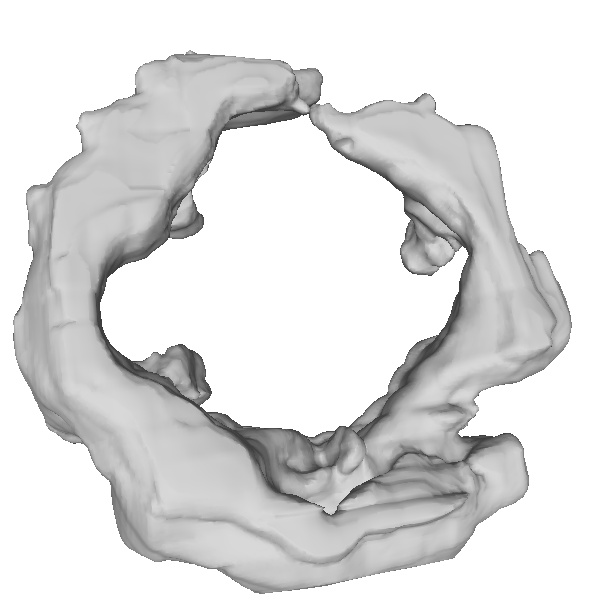} &
\includegraphics[width=.085\linewidth]{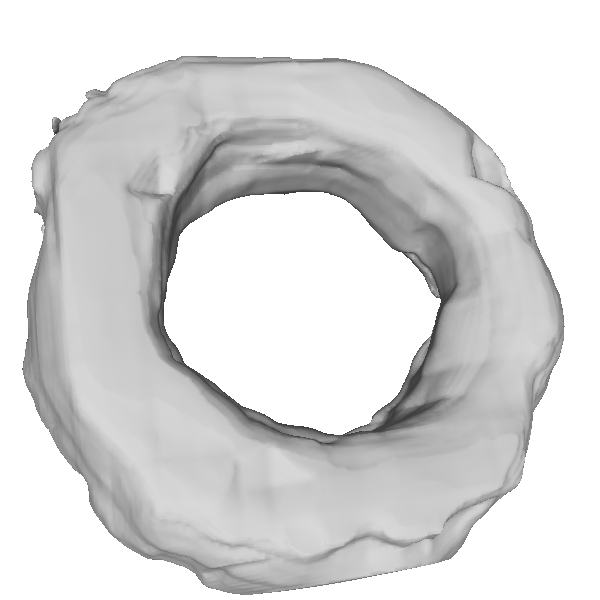} &
\includegraphics[width=.085\linewidth]{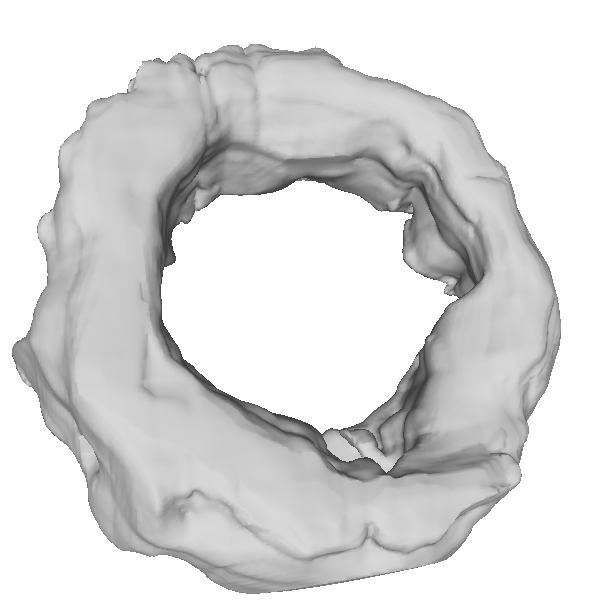} &
\includegraphics[width=.085\linewidth]{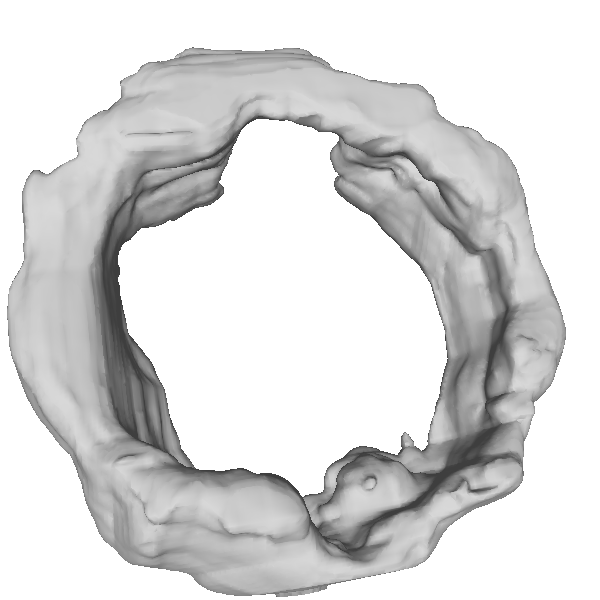} &
\includegraphics[width=.085\linewidth]{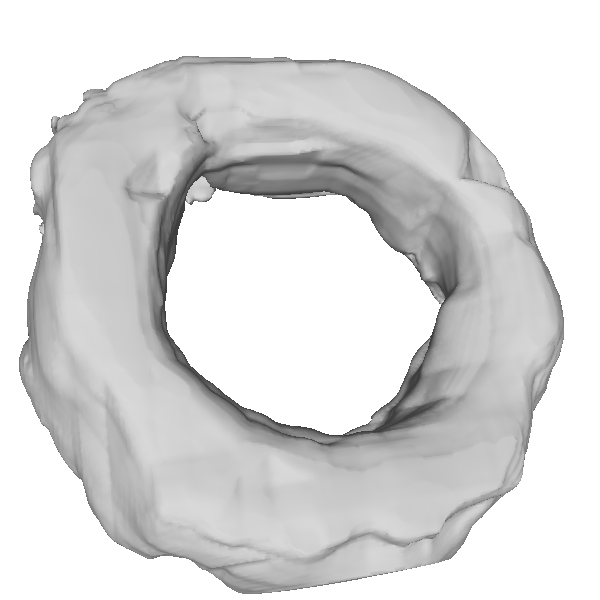} &
\includegraphics[width=.085\linewidth]{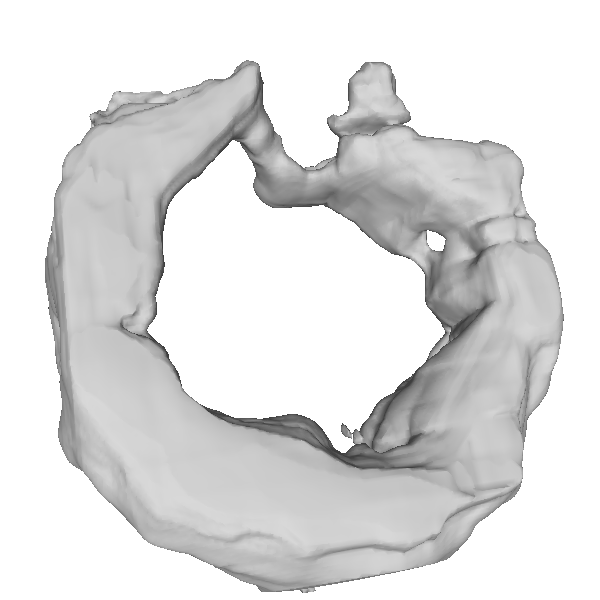} &
\includegraphics[width=.085\linewidth]{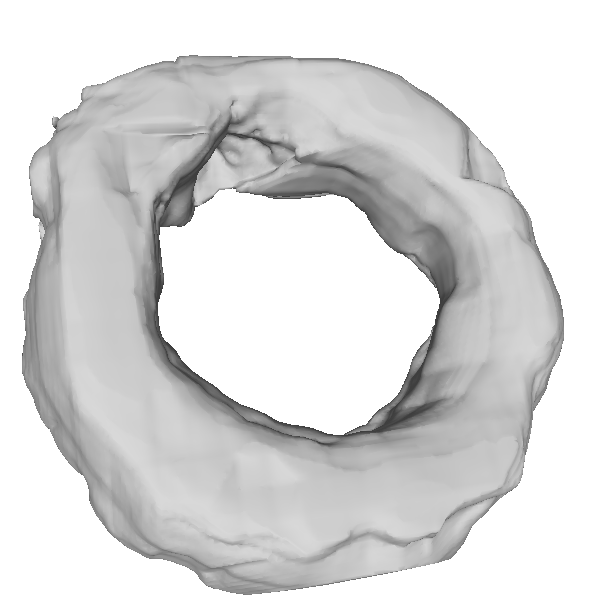} &
\includegraphics[width=.085\linewidth]{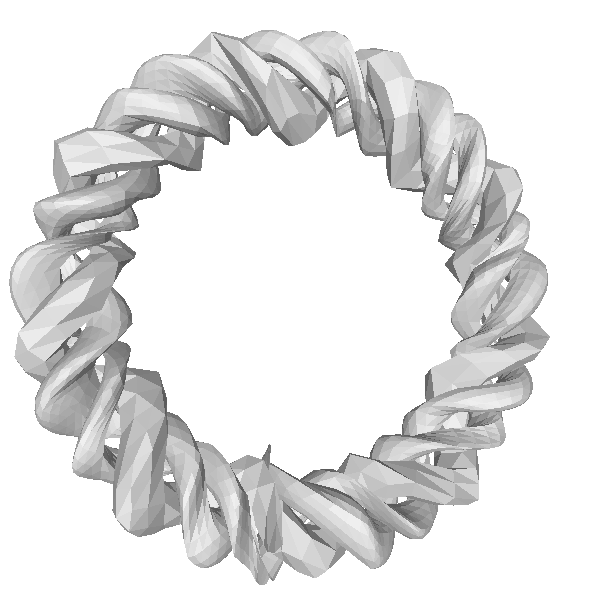} \\
\hline
{\begin{sideways} \centering \small{grid} \end{sideways}} &
\includegraphics[width=.085\linewidth]{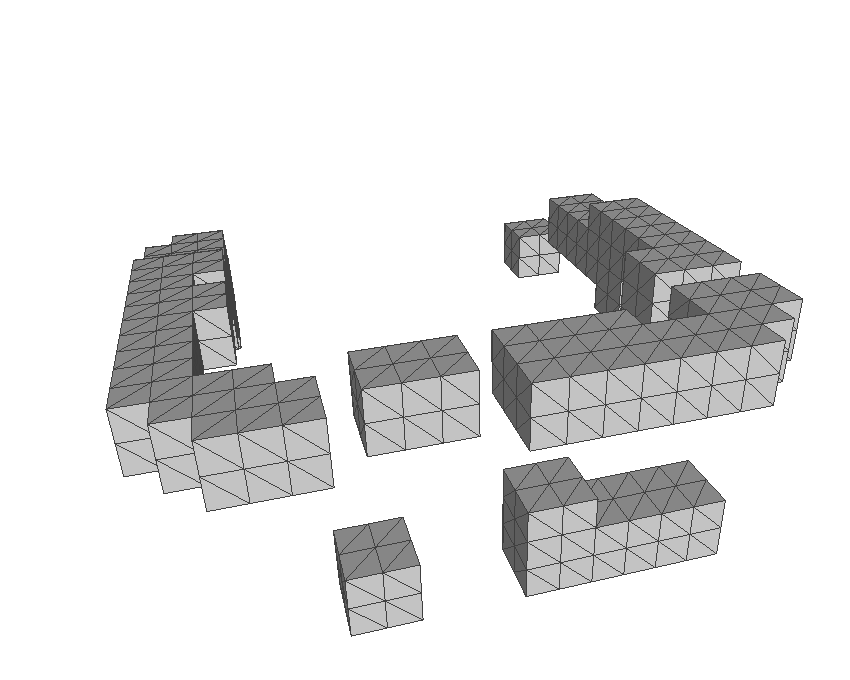} &
\includegraphics[width=.085\linewidth]{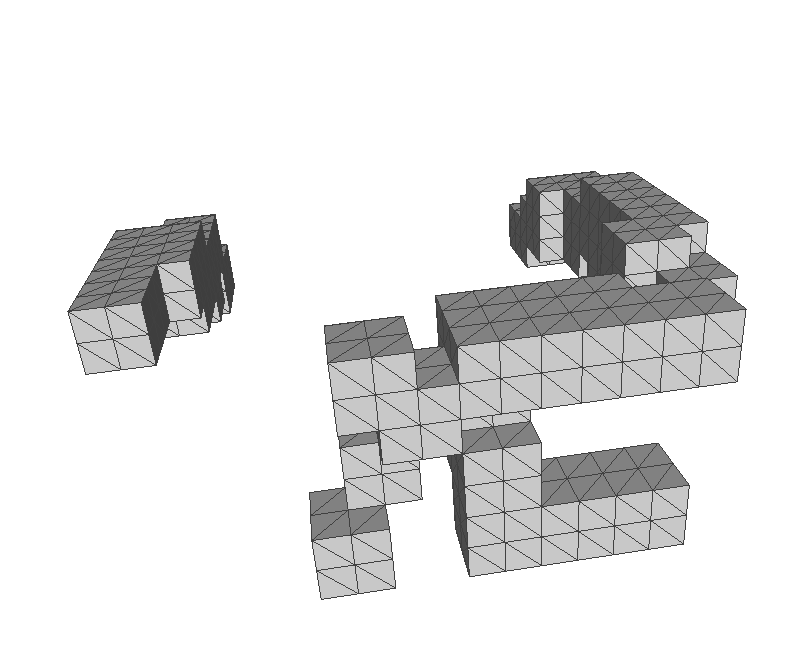} &
\includegraphics[width=.085\linewidth]{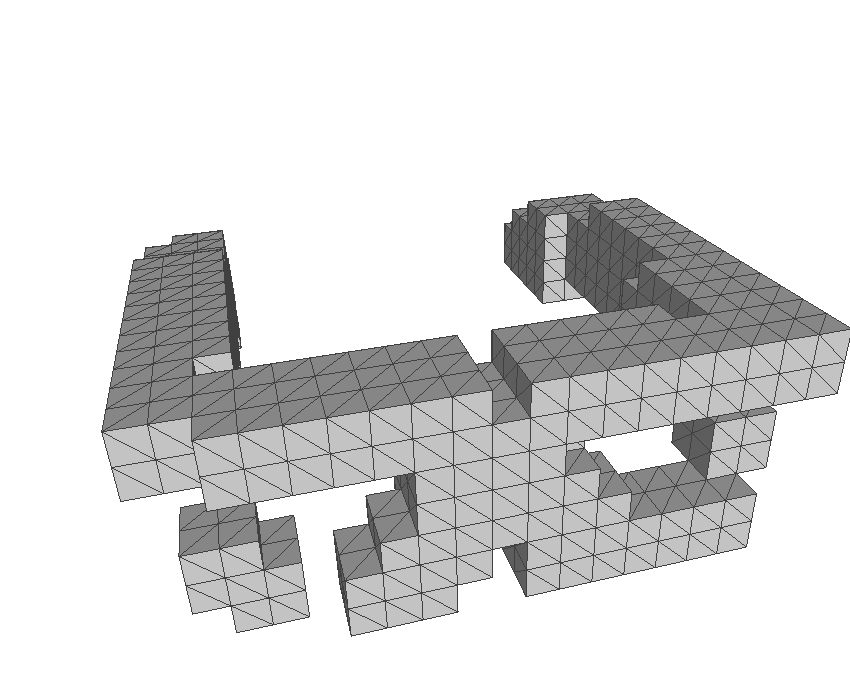} &
\includegraphics[width=.085\linewidth]{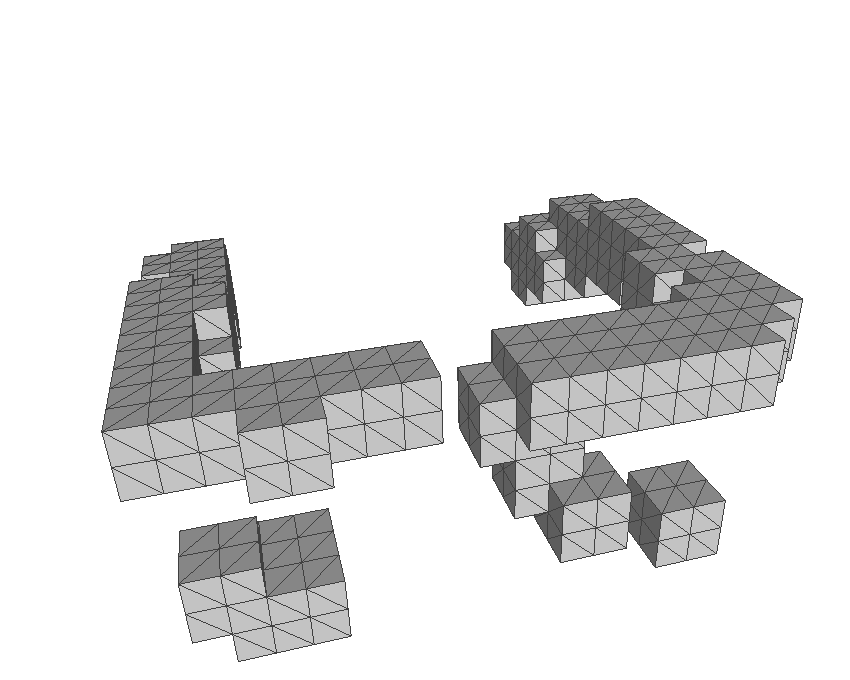} &
\includegraphics[width=.085\linewidth]{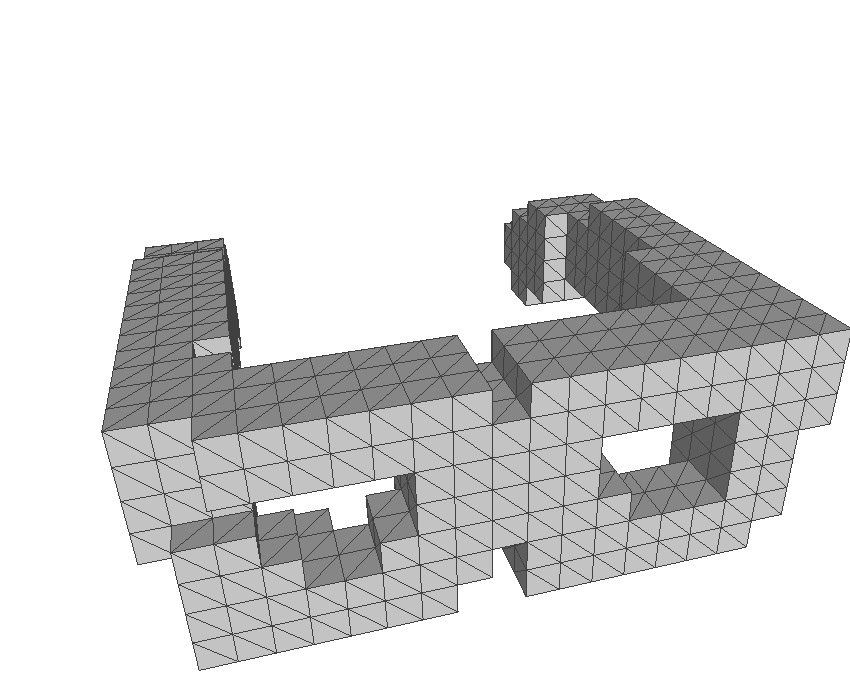} &
\includegraphics[width=.085\linewidth]{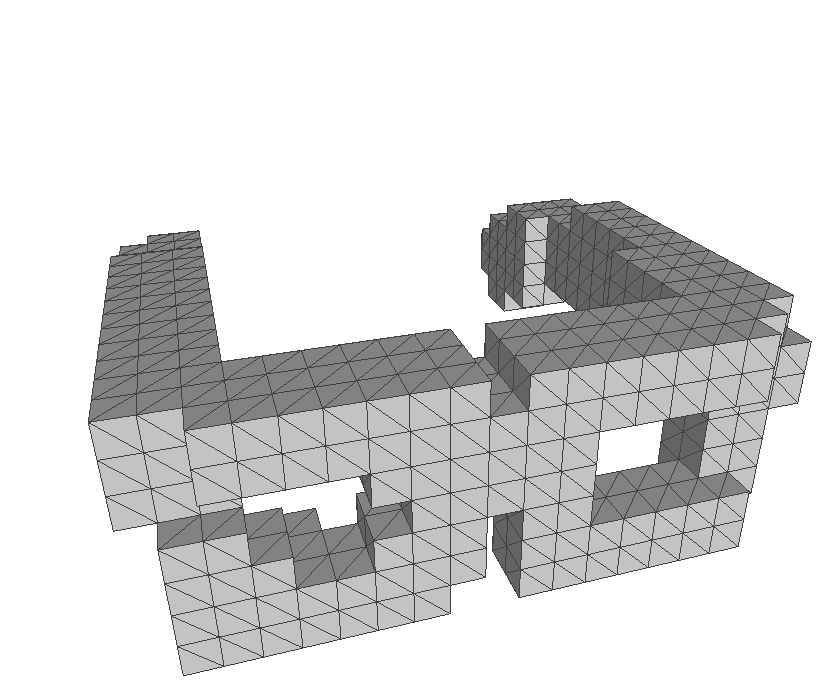} &
\includegraphics[width=.085\linewidth]{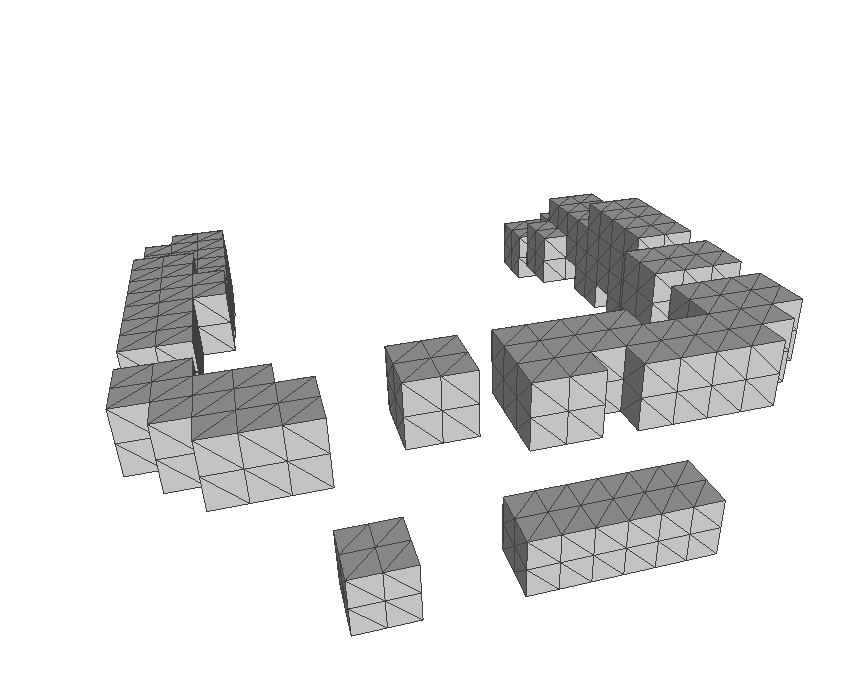} &
\includegraphics[width=.085\linewidth]{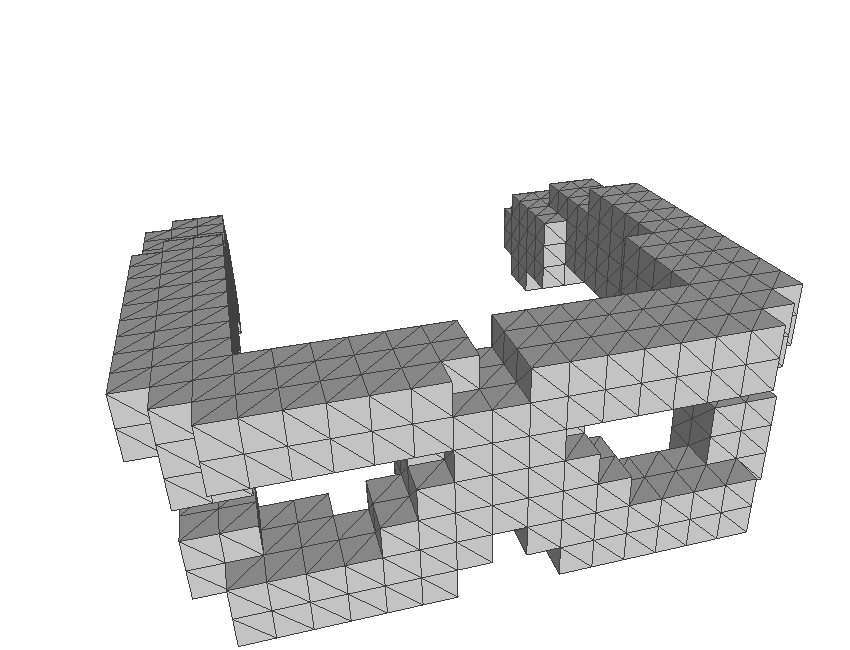} &
\includegraphics[width=.085\linewidth]{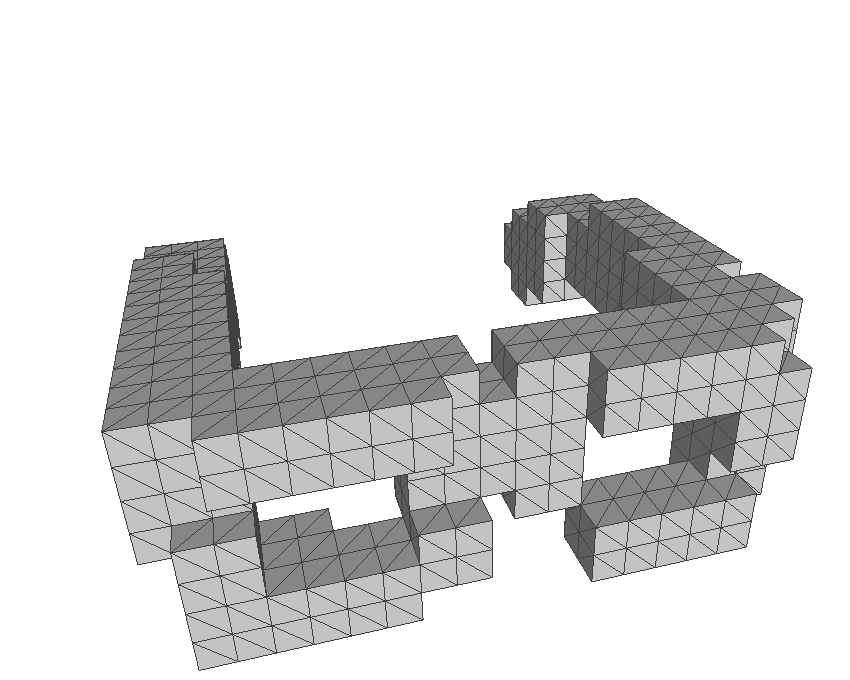} &
\includegraphics[width=.085\linewidth]{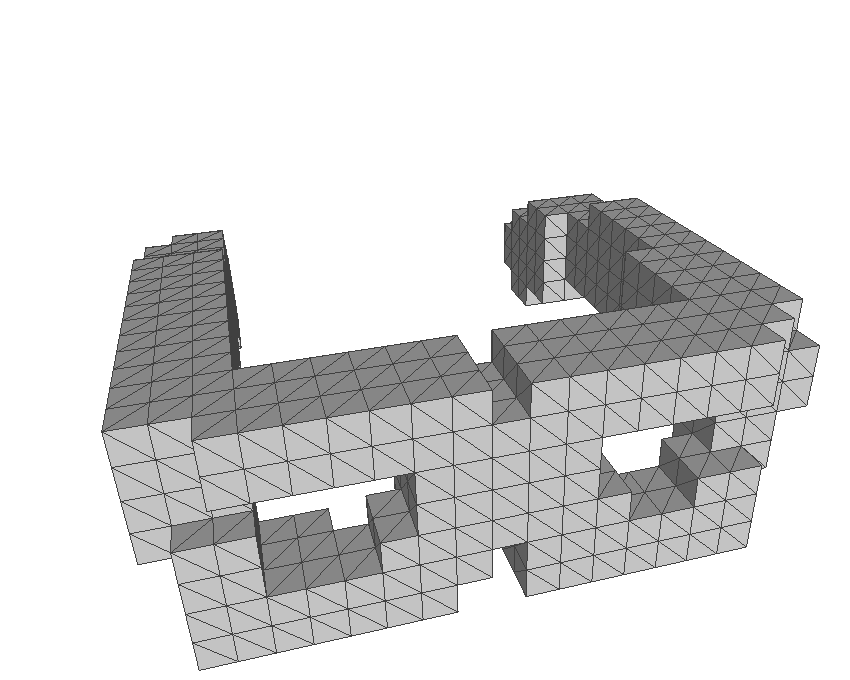} &
\includegraphics[width=.085\linewidth]{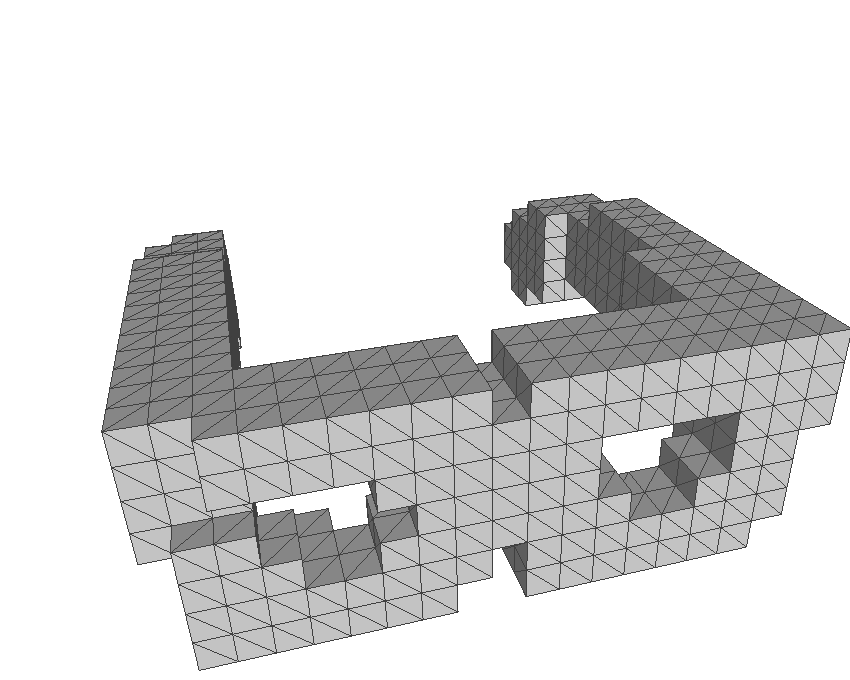} \\
{\begin{sideways} \centering \small{mesh} \end{sideways}} &
\includegraphics[width=.085\linewidth]{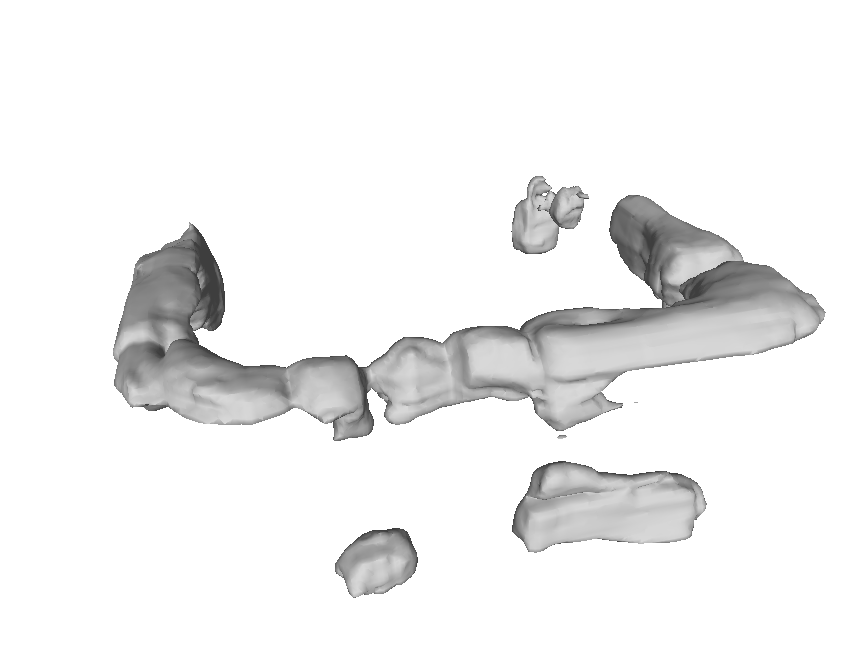} &
\includegraphics[width=.085\linewidth]{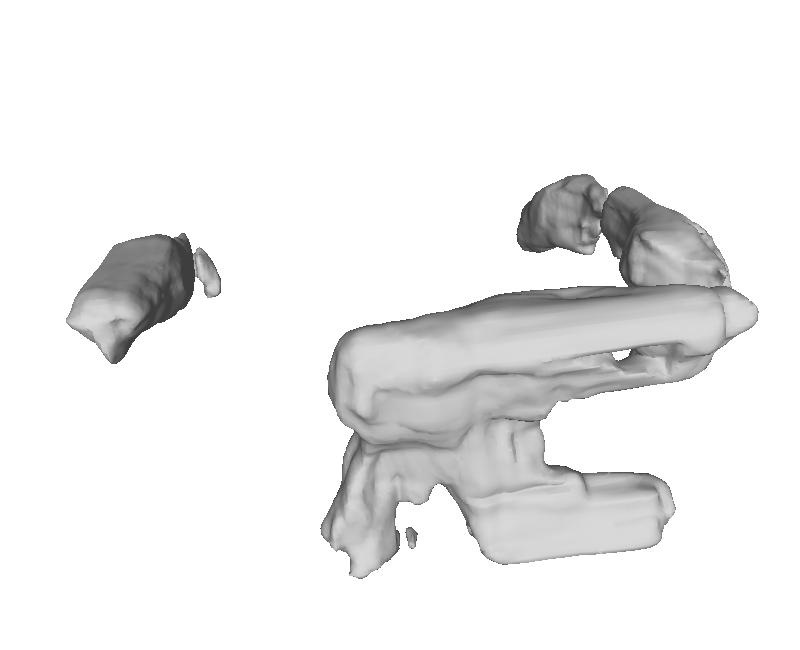} &
\includegraphics[width=.085\linewidth]{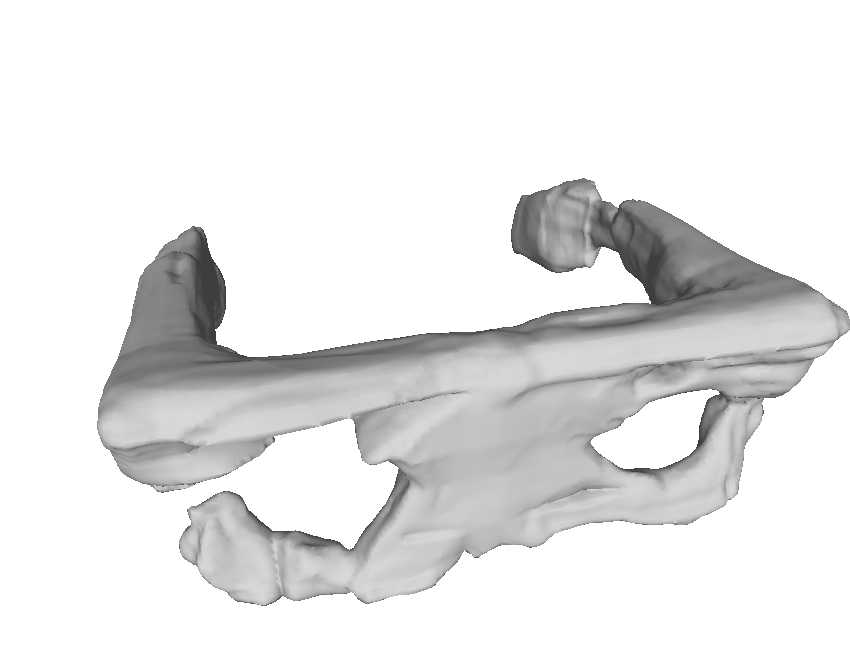} &
\includegraphics[width=.085\linewidth]{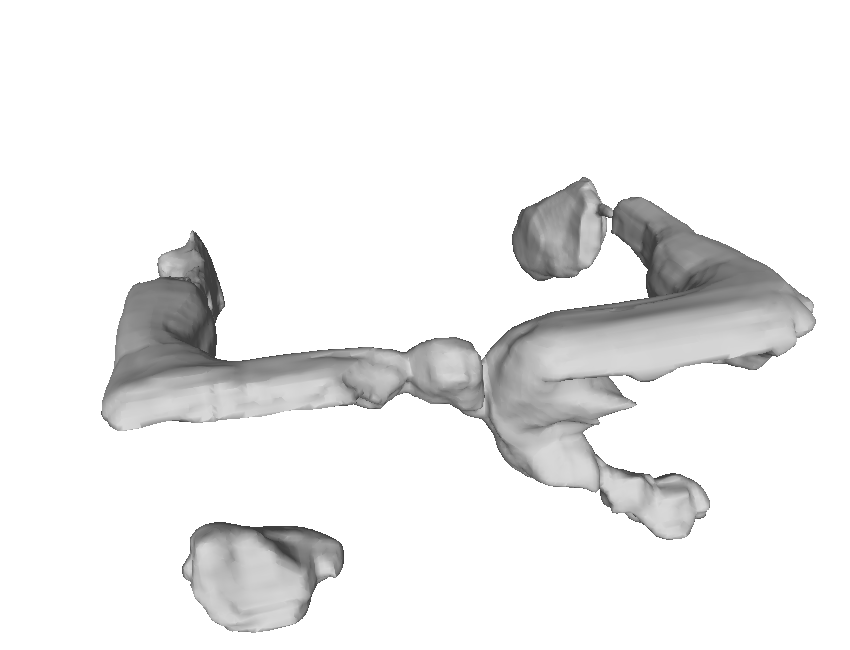} &
\includegraphics[width=.085\linewidth]{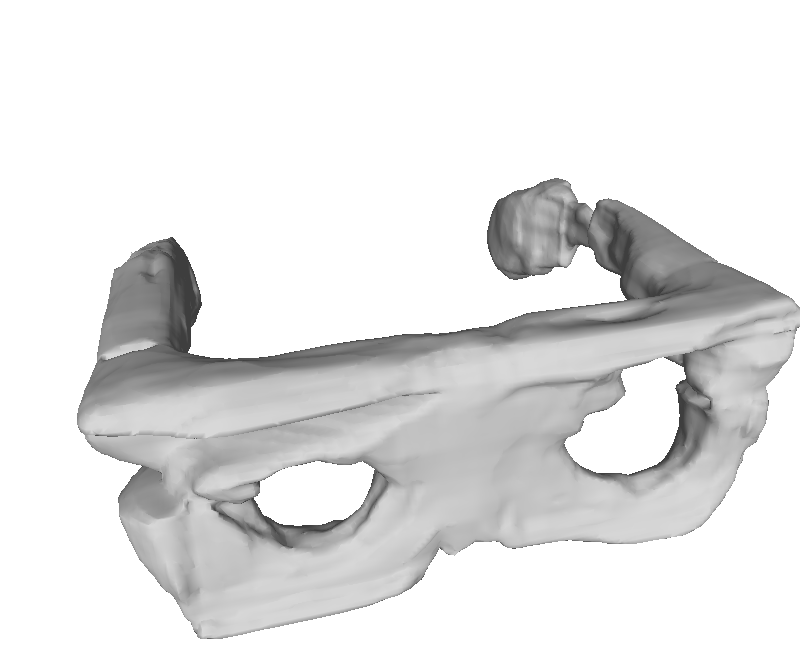} &
\includegraphics[width=.085\linewidth]{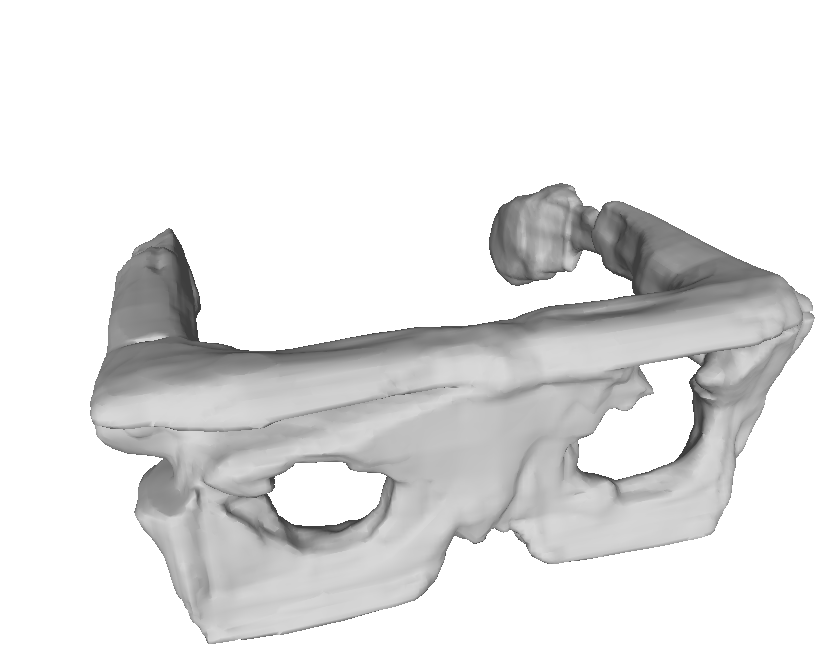} &
\includegraphics[width=.085\linewidth]{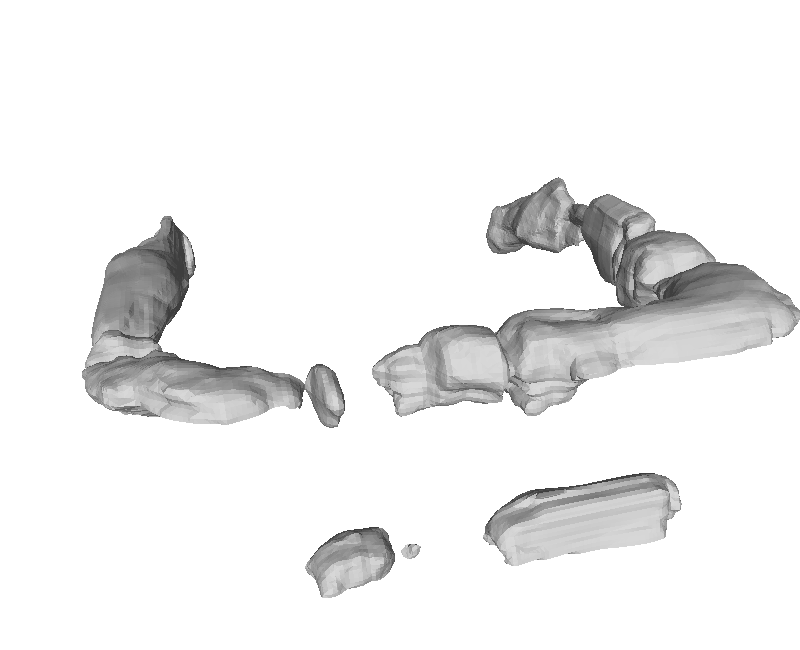} &
\includegraphics[width=.085\linewidth]{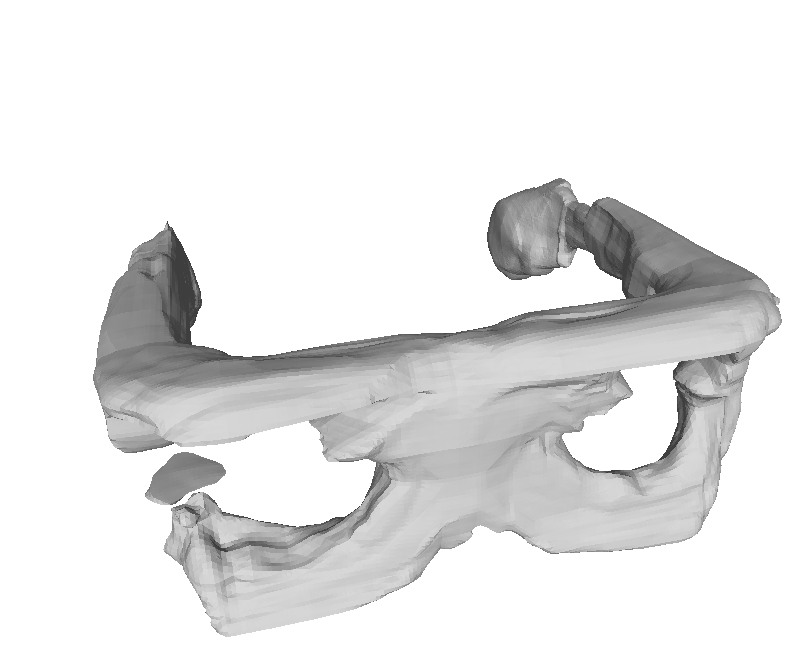} &
\includegraphics[width=.085\linewidth]{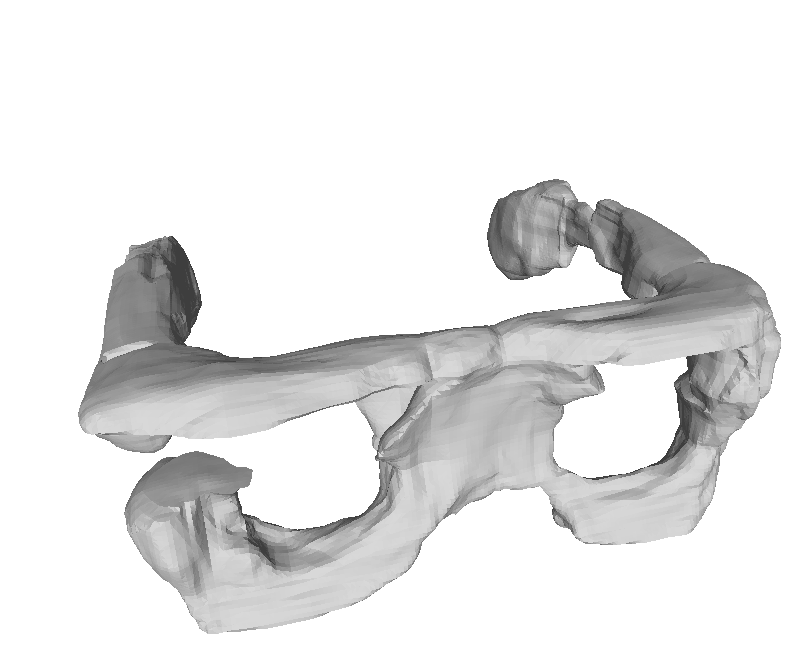} &
\includegraphics[width=.085\linewidth]{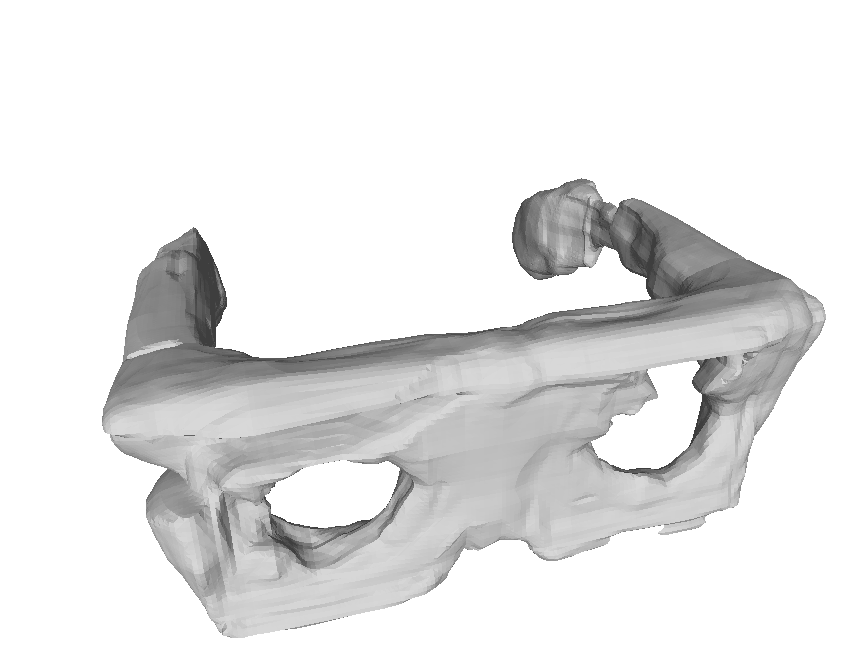} &
\includegraphics[width=.085\linewidth]{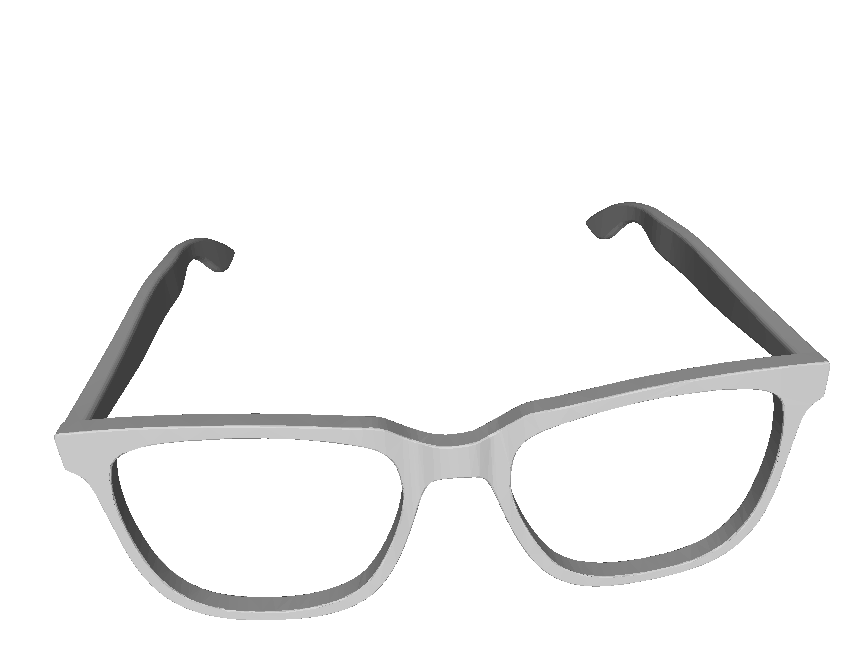} \\

\end{tabular}
\caption{Qualitative results of 3D Reconstruction on randomly selected objects from ContactDB. From left to right are: Random(R), Heuristic(H), Ours(-Coverage), Ours(-Curiosity), Ours(Knn), Ours(\#Points), Ours(\#Contact), Ours(Disagreement), Ours(Chamfer), Ours(O), Ground Truth(GT).}
\label{fig:main_results}
\end{table}
\subsection{Ablations} To better understand various design decisions, we perform extensive ablation studies investigating different component of our method:

\begin{itemize}
\item \textit{Ours(-Coverage)}: Our method trained without environment coverage reward, the other parts remain unchanged. This ablation shows the value of environment coverage reward. 
\item \textit{Ours(-Discovery)}: Our method trained without novel parts discovery reward, the other parts remain unchanged. This ablation shows the value of novel parts discovery reward. 
\item \textit{Ours(\# Points)}: This ablation replace the discovery reward with new contact points reward.
\item \textit{Ours(\# Contact)}: This ablation only replace the discovery reward with a binary score (\textit{i.e.} 1 if the hand interact with the object, 0 if not) at each timestep.
\item \textit{Ours(Knn)}: We replace the discovery reward with Knn reward. Intuitively, we want the agent to explore the most unfamiliar part of the object by maximizing the distance of new contact points with known parts. Thus, we use the mean distance of 5-nns as the reward.
\item \textit{Ours(Disagreement)}: We replace the discovery reward with a disagreement reward, where we incentive the agent to maximum the disagreement of the prediction an object after each touch. We use Alpha Shape~\cite{edelsbrunner1983shape} to predict the geometry of an object from partial observable contact points and the disagreement is measured by Chamfer distance. 
\item \textit{Ours(Chamfer)}: Our method can also be trained using a supervised exploration reward, where we incentive the agent to have a better prediction of the object after each timestep. Thus, we replace the discovery reward with an inverse chamfer distance between accumulated point cloud and ground truth. 
\end{itemize}

The results are shown in Table~\ref{table:ablation}. Comparing the first two row we note that the curiosity reward is more important to the performance than coverage reward, which validates that our method is conditioned on the partial reconstructed geometry of an object other than aggressively explore the entire state space. Ours (\# Contact) performs the worst as its a sparse and extremely noisy reward measure. By increasing the granularity of the reward, Ours (\# Points) performs slightly better than Ours (\# Contact) but is still not comparable with other ablations since it ignores object geometry information. Ours(Knn) and Ours(Disagreement) achieve similar performance as ours, which shows both Knn and Disagreement rewards encode the geometry information properly. This also proves the robustness of our method, \textit{i.e.}, it is not sensitive to the accurate value of the rewards as long as they encode rich information of the objects. 

\subsection{Qualitative Analysis}

To better understand the behavior of our method, we visualize the occupancy grid post exploration phase and final mesh post reconstruction phase for all variations of our method and all baseline methods. As shown in Figure~\ref{fig:main_results}, our method achieves high-fidelity reconstruction of convex objects. \textit{e.g.} the button of game controller (Row 4, Row 5), the clock hands of alarm clock (Row 3, Row 4). If considered the low resolution of object representation, our policy is nearly optimal, \textit{i.e.} the occupancy grid is almost the same as ground truth (Column 10, Column 11). 

Our method can also interact with the non-convex objects efficiently. \textit{e.g.} Our method successfully discovered the holes of a donut (Row 9, Row 10), an eyeglasses (Row 11, Row 12). The successful reconstruction of the cup (Row 1, Row 2) further proves that our method is able to explore the internal space of an object. Because the wall of the cup is three-grid, if an agent only explores the outer surface, the thickness information cannot be grasped. 

These qualitative results are consistent with the quantitative results. It worth noticing that Ours(\#Contact) performs very similar as random policy. This is because the working space of the dexterous hand is only 8 times the object volume, which make it easy for the robot to touch the object. And a random action is likely to lead to a positive reward of Ours(\#Contact). Thus, these two policies are likely to have similar behavior.

\begin{figure}
\centering
\subfloat{\includegraphics[width=.1\linewidth]{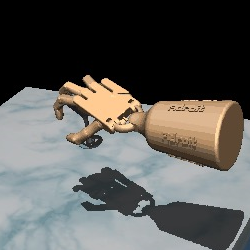}}
\subfloat{\includegraphics[width=.1\linewidth]{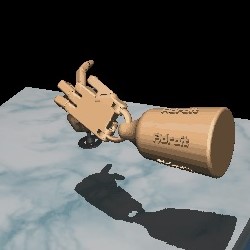}}
\subfloat{\includegraphics[width=.1\linewidth]{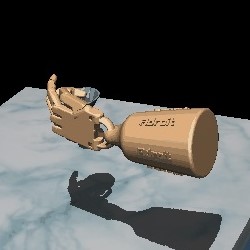}}
\subfloat{\includegraphics[width=.1\linewidth]{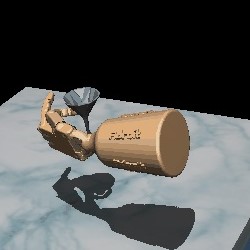}}
\subfloat{\includegraphics[width=.1\linewidth]{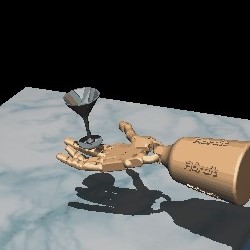}}
\subfloat{\includegraphics[width=.1\linewidth]{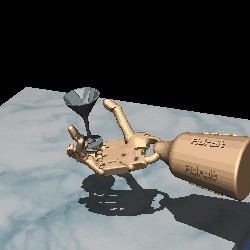}}
\subfloat{\includegraphics[width=.1\linewidth]{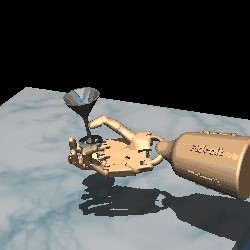}}
\\
\subfloat{\includegraphics[width=.1\linewidth]{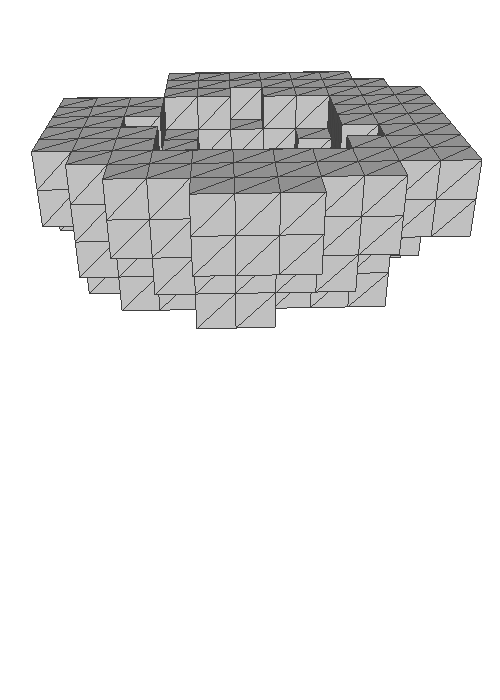}}
\subfloat{\includegraphics[width=.1\linewidth]{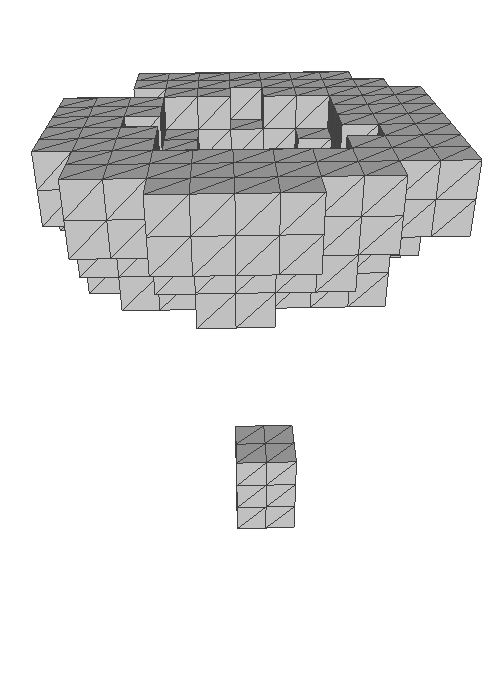}}
\subfloat{\includegraphics[width=.1\linewidth]{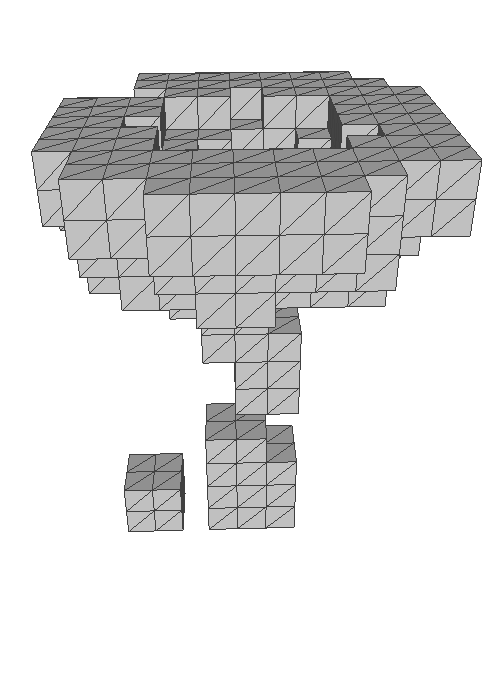}}
\subfloat{\includegraphics[width=.1\linewidth]{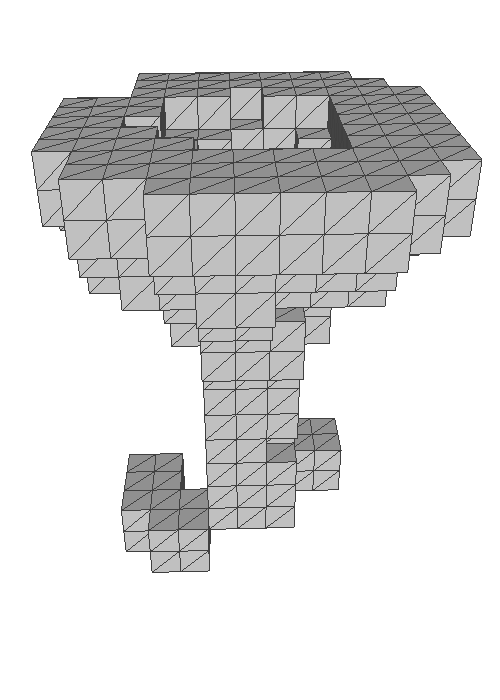}}
\subfloat{\includegraphics[width=.1\linewidth]{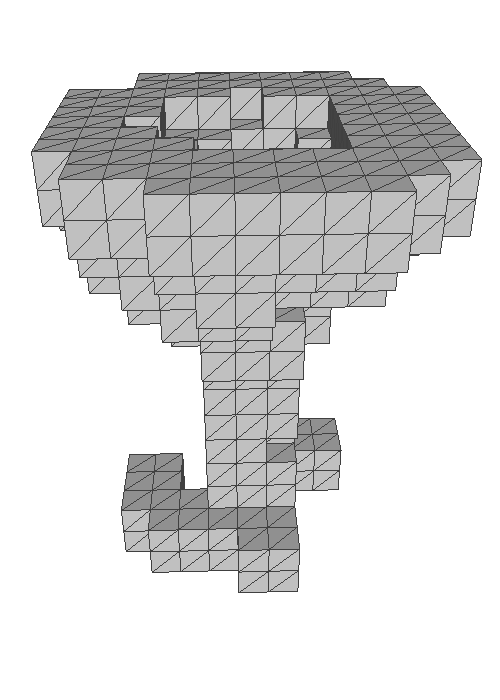}}
\subfloat{\includegraphics[width=.1\linewidth]{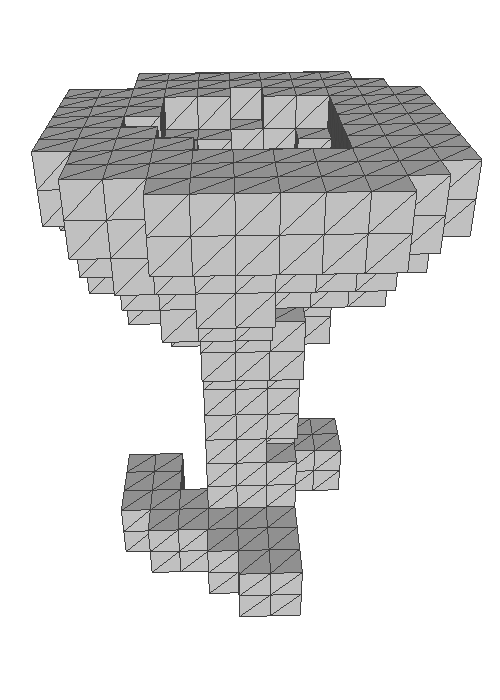}}
\subfloat{\includegraphics[width=.1\linewidth]{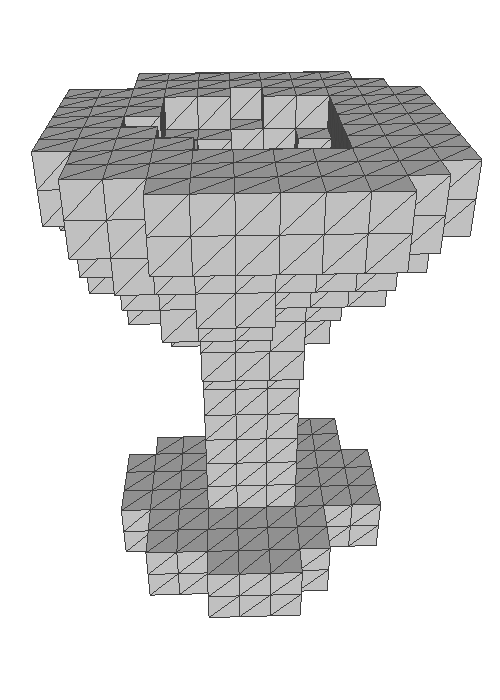}}

\caption{TOP: A sample trajectory for an agent grasping a Martini glass. BOTTOM: Progression of voxel occupancy}
\label{fig:trajectory}
\end{figure}

We further visualize the trajectory of an agent grasping a Martini glass. As shown in Figure~\ref{fig:trajectory}, our method can continually explore the unknown part of the object. For more visualization details, please refer to Supplemental Materials, or \url{https://sites.google.com/view/tslam}.



%% file: sections/6-conclusion.tex
\section{Conclusion}
In this work, we introduce \methodname~, which prepares an curiosity driven agent to exhibit effective information seeking behavior and use implicit understanding of common household items to reconstruct the geometric details of the object under exploration. Experiments demonstrate that \methodname~ is highly effective in reconstructing unknown objects of varying complexities (including non convex objects and objects with large voids) with 6 seconds of interactions.

\section{Future Works}
\label{sec: future_works}
\methodname~ while effective has a few limitations that we hope to address in followup work. Depending on the shape and inertial properties, household objects can move during interaction. Force readings of the tactile sensors can be leveraged to ensure delicate exploration of the object. In this work we focused on rigid objects. Force reading can in principle be used to adapt \methodname~ to interactively explore deformable objects as well. Surface registration of the contacts events, however, will get extremely difficult with moving surfaces. \methodname~ uses a sequential two phase approach of exploration followed by reconstruction. It can be extended to iterative improve both phases leveraging one to improve the other. Finally we aspire to close the loop by demonstrating effective manipulation of objects using reconstructions acquired via \methodname~.